\pdfoutput=1
\documentclass[11pt]{article}

\usepackage[final]{acl}

\usepackage{times}
\usepackage{latexsym}

\usepackage[T1]{fontenc}
\usepackage[utf8]{inputenc}

\usepackage{microtype}

\usepackage{inconsolata}

\usepackage{graphicx}
\usepackage{amsmath} 
\usepackage{bm}
\usepackage{booktabs}
\usepackage{multirow}
\usepackage[normalem]{ulem}
\useunder{\uline}{\ul}{}
\usepackage{color}
\usepackage{xcolor}
\definecolor{lightgray}{rgb}{0.6,0.6,0.6} % 定义灰色
\usepackage{amsfonts, amssymb}

\title{Finedeep: Mitigating Sparse Activation in Dense LLMs via \\
Multi-Layer Fine-Grained Experts} % TODO 改成 Finedeep: Mitigating Sparse Activation in Dense LLMs with a Deep Multi-Layered Fine-Grained Expert Architecture更好

\author{
    Leiyu Pan\textsuperscript{\rm 1}\footnotemark[1], 
    Zhenpeng Su\textsuperscript{\rm 2}\footnotemark[1], 
    Minxuan Lv\textsuperscript{\rm 2},
    Yizhe Xiong\textsuperscript{\rm 4},
    Xiangwen Zhang\textsuperscript{\rm 3}\\ 
    \textbf{Zijia Lin}\textsuperscript{\rm 3,4}\footnotemark[2],
    \textbf{Hui Chen}\textsuperscript{\rm 4}, 
    \textbf{Jungong Han}\textsuperscript{\rm 4},
    \textbf{Guiguang Ding}\textsuperscript{\rm 4}\\
    \textbf{Cheng Luo}\textsuperscript{\rm 3},
    \textbf{Di Zhang}\textsuperscript{\rm 3},
    \textbf{Kun Gai}\textsuperscript{\rm 3},
    \textbf{Deyi Xiong}\textsuperscript{\rm 1}\footnotemark[2] \\
    \textsuperscript{\rm 1}College of Intelligence and Computing, Tianjin University\\
    \textsuperscript{\rm 2}Chinese Academy of Sciences,
    \textsuperscript{\rm 3}Kuaishou Technology,
    \textsuperscript{\rm 4}Tsinghua University \\
    \texttt{\{lypan, dyxiong\}@tju.edu.cn} \\
    \texttt{\{suzhenpeng, linzijia\}@kuaishou.com}\\ 
}

\begin{document}
\maketitle

\renewcommand{\thefootnote}{\fnsymbol{footnote}} %将脚注符号设置为fnsymbol类型，即特殊符号表示
\footnotetext[1]{These authors contributed equally to this work.} %对应脚注[1]
\footnotetext[2]{Corresponding authors.} %对应脚注[2]
\renewcommand{\thefootnote}{\arabic{footnote}}

\begin{abstract}

% 大语言模型已经被发现能够在广泛的任务上表现出出色的性能。然而，我们发现主流的dense模型受到稀疏激活问题的困扰，这种问题限制了模型的参数利用效率，基于此，我们认为现有的dense模型还有进一步的提升空间。为了一定程度上缓解dense模型的稀疏激活问题，我们提出了Finedeep，一种层级排列的dense结构。它将传统dense模型的前馈神经网络层切分为小专家，并对这些专家进行多子层排列，使用新颖的路由方式决定不同专家的贡献程度。我们在多个尺寸的模型设置上进行了广泛的实验。我们的方法在参数量和浮点运算数与传统dense结构基本持平的情况下，PPL和benchmark结果显著优于传统的dense结构。进一步的，我们探究发现我们的方法在深度和宽度，即专家排列子层数和每个子层排列的专家数达到平衡时达到最优效果。我们还通过消融实验证明排列多个子层和子层内排列多个专家的必要性，并实证性地证明了我们提出的方法确实能够缓解稀疏激活现象。
Large language models have demonstrated exceptional performance across a wide range of tasks. However, dense models usually suffer from sparse activation, where many activation values tend towards zero (i.e., being inactivated). We argue that this could restrict the efficient exploration of model representation space. To mitigate this issue, we propose \textbf{Finedeep}, a \textbf{deep}-layered \textbf{fine}-grained expert architecture for dense models. Our framework partitions the feed-forward neural network layers of traditional dense models into small experts, arranges them across multiple sub-layers. A novel routing mechanism is proposed to determine each expert's contribution. We conduct extensive experiments across various model sizes, demonstrating that our approach significantly outperforms traditional dense architectures in terms of perplexity and benchmark performance while maintaining a comparable number of parameters and floating-point operations. Moreover, we find that Finedeep achieves optimal results when balancing depth and width, specifically by adjusting the number of expert sub-layers and the number of experts per sub-layer. Empirical results confirm that Finedeep effectively alleviates sparse activation and efficiently utilizes representation capacity in dense models.

\end{abstract}

\section{Introduction}

% 不同模型激活函数输出分布，所选模型的激活函数均为silu
% 横轴表示激活值，纵轴表示激活值的分布密度，即不同模型在不同激活值上的频率分布。
\begin{figure}[t]
\centering
\centerline{\includegraphics[scale=0.32]{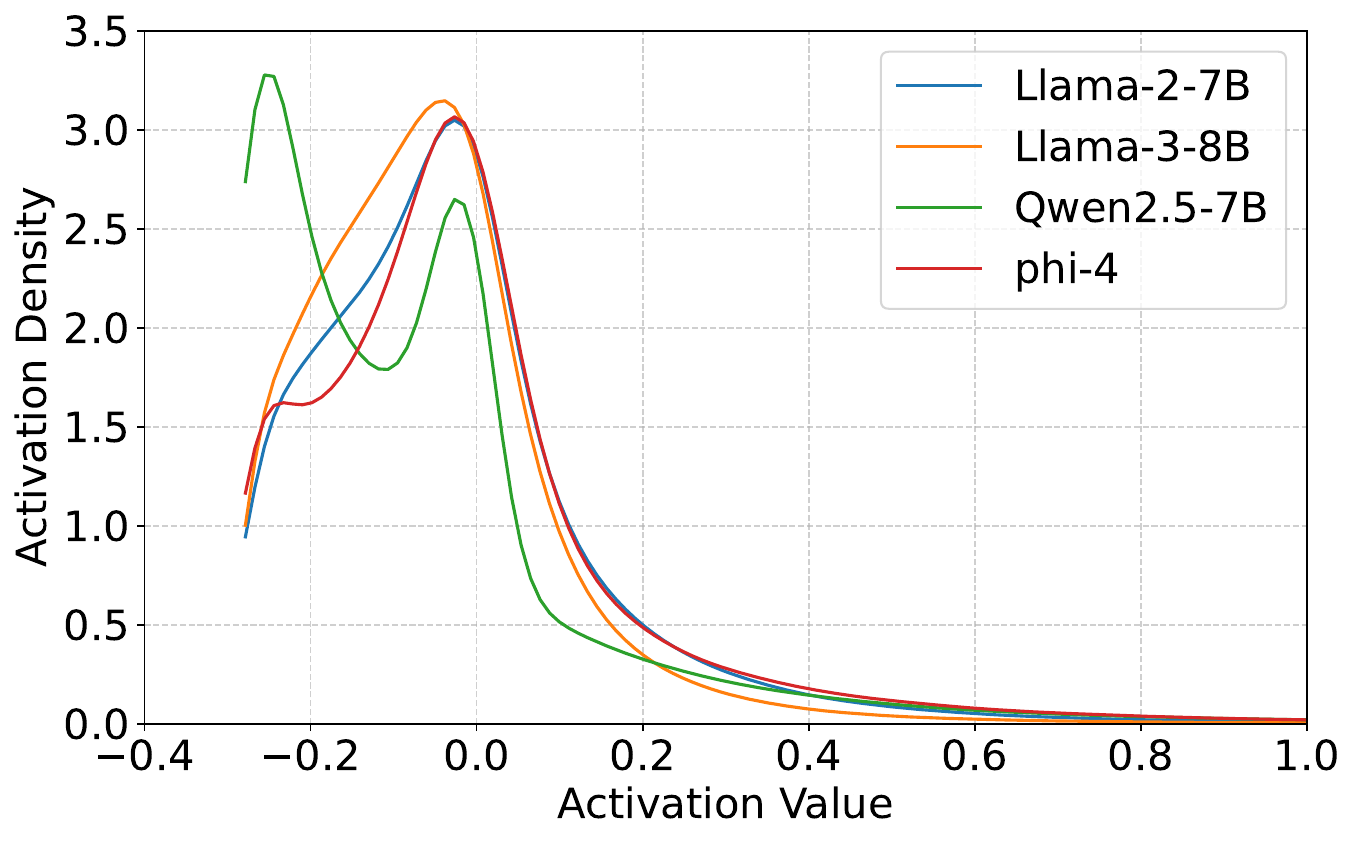}}
\caption{Distribution of activation function outputs across various models \cite{touvron2023llama, dubey2024llama, yang2024qwen2, abdin2024phi}, where all selected models use the SiLU activation function. The horizontal axis represents the activation values, while the vertical axis denotes the distribution of activation values across different models.}
\label{fig1}
\end{figure}

% 近年来，大语言模型由于其在广泛领域上的出色表现而引起人们的关注，关于大模型的预训练缩放定律证明，随着大语言模型尺寸的增大，模型在下游任务上的性能会不断提升。不幸的是，这种通过扩展模型尺寸提升模型性能的手段往往带来难以承担的高额的成本。所以现有的探索往往在固定的模型参数成本下，尽可能地拓展模型性能的上限空间。
Large language models (LLMs) have recently gained much attention for their exceptional performance across various tasks \cite{achiam2023gpt, touvron2023llama, dubey2024llama, yang2024qwen2}. Scaling laws at the pre-training stage of LLMs suggest that increasing model size could consistently enhance performance on downstream tasks \cite{DBLP:journals/corr/abs-2001-08361, DBLP:journals/corr/abs-2203-15556}. However, such improvement often comes at an exorbitant computational cost. As a result, maximizing model performance within a fixed parameter budget has emerged as an efficient paradigm, aiming to push the upper bound of the model performance without significantly increasing resource demands \cite{zhang2024tinyllama}.

% 通过前期实验验证，我们发现现有的稠密模型在实际计算过程中存在稀疏激活现象，如图1所示。我们认为，这一现象可能成为提升模型性能上限的关键突破点。具体而言，稀疏激活是指模型经过激活函数处理后，其输出中大部分值趋近于0。由于这些值较小，即便与模型参数相乘，贡献依然有限，对最终结果的影响微乎其微，导致激活值的利用率较低。如果能够提高激活值的利用率，模型的表示能力将更加强大，从而能够拟合更复杂的特征，进一步拓展模型的性能上限。
Along with model scaling, recent studies disclose that dense models usually exhibit a sparse activation phenomenon during computation \cite{zhang2022moefication}, as illustrated in Figure \ref{fig1}. Specifically, sparse activation refers to the fact that most values output from the activation functions tend to be close to zero \cite{lilazy, luo2024sparsing}. Since these small values contribute marginally when multiplied by model parameters, their impact on the final output remains limited, leading to inefficient activation utilization. We argue that addressing sparse activation could serve as a new channel for further improving the upper limit of model performance. By improving the effective utilization of activation values, we could enhance the representational capacity of models, enabling them to capture additional complex features.

% 受到XMOE工作的启发，我们的方法，Finedeep，通过将传统dense模型的FFN切分成细粒度的专家来缓解稀疏激活的现象。特别的，我们没有采取以往moe专家单层排列的方式，而是采取了一种包含更多表示空间的多层排列专家的方式，从拓展模型的宽度转而拓展模型的深度。另外我们也采取了一种新颖的串联的路由策略，我们利用每层排列的专家的输出经过router计算路由分数，然后进行软加权求和操作，这与其他利用专家输入来计算路由分数，再将路由分数作为专家权重的并联策略不同。特别的，软加权求和操作往往会受到不同专家之间竞争的影响，因为softmax归一化会强制形成少数专家占主导地位的概率分布，从而导致贡献不平衡。为了缓解这一问题，我们借鉴了XX的见解，用sigmoid函数取代了传统的softmax函数用于路由分数归一化。

% 与之前采取单层排列专家的方式的moe不同，我们采取多层排列专家的方式。每一层的参数总量更少，但是参数激活幅度更高。这样可以提供更广泛的表示空间。
Following this direction, we hence propose Finedeep, a new dense architecture with deep-layered fine-grained experts to mitigate sparse activation. Our framework partitions the feed-forward networks (FFNs) of a traditional dense model into fine-grained experts. Unlike previous Mixture-of-Experts (MoE) architectures that employ a single-layer expert arrangement, we adopt a multi-layer expert arrangement. While each layer has fewer total parameters, it achieves a higher magnitude of parameter activation. This design results in a richer representational space \cite{su2024cartesianmoe}. 

We further propose a novel routing strategy to efficiently combine depth-wise experts. Instead of computing routing scores based on expert inputs, as done in conventional approaches, we leverage expert outputs to determine routing scores. The final expert outputs are then combined using a soft-weighted summation. Previous soft-weighted summation often suffers from competition among different experts, as softmax normalization forces a probability distribution where a few experts dominate, leading to imbalanced contributions. To mitigate this, we build upon the insights from \citet{liu2024deepseek} and use the sigmoid function to replace the traditional softmax function for routing score normalization.

It is important to highlight that our approach differs from the MoE framework. In our method, all parameters are actively involved in the computational process. Our goal is to increase the activation rate of all parameters and ensure that the parameters of all experts are fully utilized, rather than activating only a subset of experts to save computation, as is done in the MoE architecture. Furthermore, our approach does not introduce additional parameters; instead, it disassembles the original FFN into fine-grained experts, avoiding the need to expand parameters and consequently increasing the total model size as seen in MoE.

% 为了验证我们提出方法的有效性，我们进行了广泛的预训练实验。通过在不同尺寸模型、不同切分专家数量的实验设置上进行广泛的实验证明，我们的方法在和传统dense模型参数量基本相同的情况下，在困惑度指标和下游benchmark任务指标上表现的更好。进一步的，我们通过不同排列子层数的超参数实验也发现，适度地加深模型网络而不是拓宽模型网络对模型性能是有帮助的。最后，我们实证性地证明了我们的方法确实能够缓解稀疏激活的问题。
% 超参数实验揭示了当宽度和深度达到平衡时达到最佳效果。
To validate the effectiveness of Finedeep, we conduct extensive LLM pre-training experiments. Through experiments across different model sizes and varying numbers of fine-grained experts, we demonstrate that Finedeep consistently outperforms traditional dense models in both perplexity (PPL) and downstream benchmarks, while maintaining a comparable number of parameters. Furthermore, hyper-parameter studies reveal that optimal results are achieved when width and depth are balanced. Finally, our empirical analysis confirms that Finedeep effectively mitigates sparse activation, enhancing overall model representation capacity.

% 我们的贡献可以归纳如下：
% 为了应对稠密模型中出现的稀疏激活的现象，我们提出了Finedeep，通过将dense模型的FFN层切分为细粒度专家缓解问题。
% 我们的方法引入了新颖的专家排列和路由策略，通过在一定程度上稳定地加深网络提高模型性能。
% 我们在广泛的实验上证明了Finedeep相比传统dense模型的优势，并实证性地证明了方法确实可以缓解稠密模型稀疏激活的问题。
The main contributions of our work are summarized as follows.
\begin{itemize}
  \item To address the issue of sparse activation in dense models, we propose Finedeep, which partitions FFNs in dense models into fine-grained experts.
  \item Our approach introduces innovative expert arrangements and routing strategies, enhancing model performance while improving the stability of deep networks.
  \item Through extensive experiments in LLM pre-training, we demonstrate the superiority of Finedeep over traditional dense models and empirically validate its ability to alleviate sparse activation in dense models.
\end{itemize}

% TODO introduction 的问题，主要是还没说清楚跟MoE或者是CartesianMoE的差别，如果仅仅是routing改动，感觉还是不够distinct，这块需要再琢磨和polish一下。另外就是如何结合稀疏激活这个问题来设计你的结构，比方说，既然是稀疏激活，是否我们可以考虑将参数分配到两层，这样可以每一层都少一点参数，但是激活幅度更高一些，提高网络参数利用率，类似于这样的视角。我看related work最后一段其实写得还不错，可以在introduction这里先写，然后related work再提一下，其实就是说明我们这里是希望充分提高参数的激活率，希望所有expert都能被充分激活，而不是像MoE那样是为了节省计算，另外跟MoE的不同也体现在我们不扩充参数，而是把原来的FFN直接拆解成finegrained expert，而不是像MoE那样扩展参数。

% 需要注意的是，我们的工作和moe有不同之处。在我们的方法中，所有的参数都同时参与计算过程。我们希望提高所有参数的激活率，希望所有的专家的参数都能够被充分激活，而不是像moe架构为了节省计算而只激活某几个专家的参数。另外，我们的方法没有引入额外的参数，直接把原有的FFN拆解为细粒度专家，而不是像moe那样扩展参数从而增大模型总参数量。

\section{Related Work}

% 子图a代表原有的稠密模型的结构，子图b代表我们提出的Finedeep模型结构，子图a和子图b的连接处代表原有稠密模型到我们提出的Finedeep模型结构的变化过程
\begin{figure*}[t]
\centering
\centerline{\includegraphics[scale=0.48]{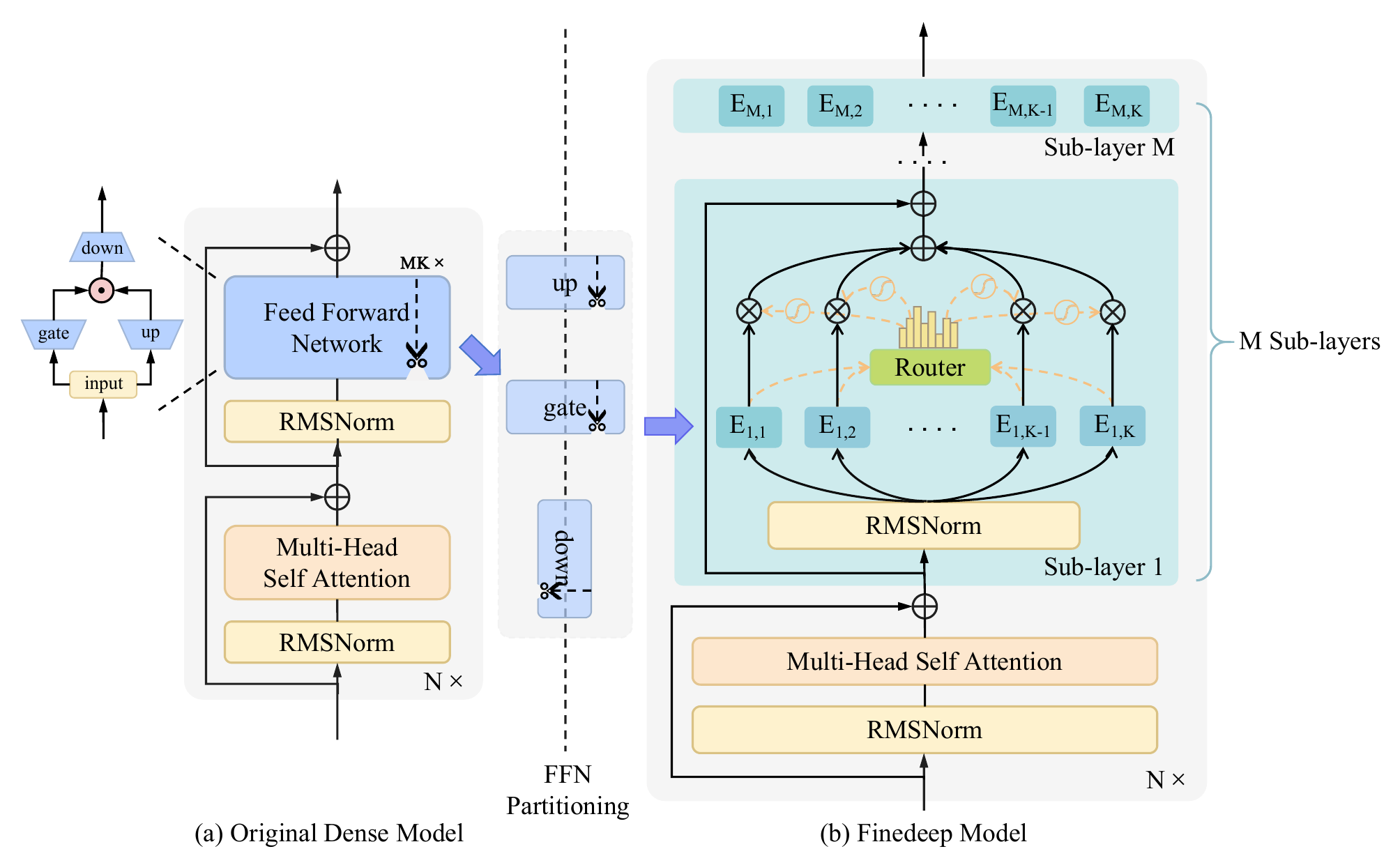}}
\caption{Illustration of the proposed Finedeep. Subfigure (a) illustrates the structure of the original dense model. Subfigure (b) demonstrates the structure of our proposed Finedeep model. Each FFN in the dense model is partitioned into $M\times K$ experts distributed along $M$ sub-layers with $K$ experts per sub-layer. The connection between subfigures (a) and (b) represents the transformation process from the original dense model to the Finedeep model.}
\label{fig2}
\end{figure*}

% 目前通用的dense模型架构都是大部分都是基于decoder-only的transformer架构，这种模型架构能够利用参数空间建模丰富的知识从而展现出良好的性能，比如模型的前馈网络层（FFN）可以理解为存储大量知识的地方，但是有研究指出，dense模型的FFN层在训练过程中存在稀疏激活的现象，即激活函数输出大部分值为较低的值，这些较低的值在后续的矩阵乘法中的贡献是非常有限的。这种现象意味着模型的激活值不能被充分利用从而导致潜在的资源浪费，并且随着训练过程的不断深入，模型稀疏激活现象越来越明显。
Current dense model architectures are predominantly based on the decoder-only transformer, which effectively leverages the parameter space to encode rich knowledge and deliver strong performance \cite{DBLP:conf/nips/BrownMRSKDNSSAA20, DBLP:journals/corr/abs-2302-13971, touvron2023llama, dubey2024llama}. FFN layers in these models are often regarded as a key component for storing substantial amounts of knowledge \cite{geva2021transformer, dai2022knowledge}. However, it has been observed that FFN layers in dense models exhibit sparse activation during the training process \cite{zhang2022moefication}, where the majority of the output values from the activation function are low, contributing marginally to subsequent matrix multiplications. This indicates that the activation values are not fully utilized, leading to a potential waste of resources. Moreover, the phenomenon of sparse activation becomes increasingly pronounced as the training process progresses \cite{luo2024sparsing}.

% 为了缓解dense模型中出现的稀疏激活问题，XX通过将dense模型转换为moe架构模型。首先确定稠密模型稀疏激活的模式，然后再根据这种模式确定划分专家的方法，最大化激活专家内参数的激活密度。XMOE在moe架构的单个专家内也发现了稀疏激活问题，它通过将专家切割为细粒度专家来缓解这种问题。与上述方法不同的是，我们的方法从始至终都是在dense模型中进行的，并没有选择moe架构通用的激活top-k专家的方法来避免稀疏激活问题，而是在所有参数参与计算的情况下减小稀疏激活问题。
To address the sparse activation issue in dense models, \citet{zhang2022moefication} transforms the dense model into a MoE architecture. First, the pattern of sparse activation in the dense model is identified, and then experts are partitioned based on this pattern to maximize the activation density within each expert. \citet{yang2024xmoe} also identifies the sparse activation problem within individual experts in the MoE architecture and mitigates this by dividing the experts into fine-grained experts. Unlike the aforementioned methods, our approach works entirely within the dense model framework. We do not adopt the common MoE strategy of activating the top-k experts to avoid sparse activation \cite{fedus2022switch, DBLP:conf/iclr/LepikhinLXCFHKS21}. Instead, we reduce sparse activation with all parameters contributing to the computation.

\section{Methodology}

% 所提出的方法Finedeep如图2所示。首先，为了应对稠密模型中会出现的稀疏激活问题，我们将传统稠密模型中的FFN层进行切分成小专家。然后我们采取新颖的专家排列和路由策略，使用多层排列专家的方式拓展模型深度，并且使用非线性路由对专家输出进行加权。上述策略使模型能够更有效地捕捉复杂特征，从而提升模型的表达能力和泛化性能。
The proposed method, Finedeep, is illustrated in Figure 2. First, to address the issue of sparse activation that commonly arises in dense models, we decompose FFNs of traditional dense models into fine-grained experts. Then, we present a novel expert arrangement and routing strategy: expanding the model depth by stacking multiple sub-layers of arranged experts and applying nonlinear routing to weight the expert outputs.

\subsection{FFN Partitioning}

% 传统的FFN层前向传播的过程包含将输入投影到中间维度，再投影回初始维度的操作。为了控制在FFN层中通过信息的比例，很多模型选择在FFN中加入门控矩阵。通常可以将FFN层的操作表示为如下公式：
In a traditional FFN layer, forward propagation involves projecting the input into an intermediate representation with a different dimension via an ``up'' projector before mapping it back to the representation with the original dimension via a ``down'' projector, as illustrated in Figure \ref{fig2}(a). Notably, most modern models employ the SiLU activation function, thus incorporating an additional gating matrix \cite{touvron2023llama}. % TODO 这个其实是swiGLU还是SiLU的激活方式，直接提一下SwiGLU或者SiLU说是最常用的，再展开这么讲更好
The operations within the FFN layer are typically represented by the following equation:

\begin{equation}
\mathrm{FFN}(\mathbf{\hat{h}}_t^l)=(\sigma (\mathbf{\hat{h}}_t^l \bm{W}_{g}) \odot \mathbf{\hat{h}}_t^l \bm{W}_\mathrm{up})\bm{W}_\mathrm{down}
\end{equation}
% 其中，Wg是门控矩阵，Wup是用于提升特征维度的投影矩阵，Wdown则用于将维度映射回原空间。Wg/Wup/Wdown分别对应fig2 子图a的FFN的gate/up/down三个矩阵。\sigma代表激活函数，$\mathbf{\hat{h}}_t^l$表示第l层多头注意力模块的输出。
where $\bm{W}_g$ is the gating matrix, $\bm{W}_\mathrm{up}$ is the projection matrix that expands the feature dimensions, while $\bm{W}_\mathrm{down}$ maps the features back to the original space. $\bm{W}_g$, $\bm{W}_\mathrm{up}$, and $\bm{W}_\mathrm{down}$ correspond to the gate, up, and down matrices of the FFN in the subfigure (a) of Figure \ref{fig2}, respectively. The function $\sigma$ represents the activation function, introducing non-linearity to enhance the model’s expressiveness. $\mathbf{\hat{h}}_t^l$ denotes the output of the Multi-Head Self Attention (MHA) module at the $l$th layer.

% 为了缓解稠密模型中出现的稀疏激活现象，我们将FFN进行分解为更小的专家。具体来说，我们首先确定切分的专家数量，即专家排列的子层数M乘以每个子层的专家数N。然后我们在中间维度上对Wg/Wup/Wdown三个矩阵进行切分，使每个专家的计算逻辑与原始FFN层保持一致，仅在中间维度上有所不同。即对于专家i来说，其计算过程表示如下：
To mitigate the sparse activation phenomenon observed in dense models, we decompose the FFN into smaller expert units. Specifically, we first determine the number of experts to be sliced, which is given by the number of sub-layers containing experts, $M$, multiplied by the number of experts per sub-layer, $K$. We then partition the three matrices $\bm{W}_g$, $\bm{W}_\mathrm{up}$ and $\bm{W}_\mathrm{down}$ along the intermediate dimensions. This ensures that the computational logic of each expert remains consistent with that of the original FFN layer, differing only in the intermediate dimensions. For a given expert $i$, the computation is as follows:

$$
\mathrm{FFN}_{i}(\mathbf{\hat{h}}_t^l)=(\sigma (\mathbf{\hat{h}}_t^l \bm{W}^{(i)}_{g}) \odot \mathbf{\hat{h}}_t^l \bm{W}^{(i)}_\mathrm{up})\bm{W}^{(i)}_\mathrm{down} $$
\begin{equation}
\text{where} \quad 1 \leq i \leq MK
\end{equation} % TODO i \in \mathbb{Z} 可不写，更简洁，不然容易遇到reviewer挑刺说没解释Z是啥
where $\bm{W}^{(i)}_{g}$, $\bm{W}^{(i)}_\mathrm{up}$ and $\bm{W}^{(i)}_\mathrm{down}$ represent the sliced weight matrices corresponding to expert $i$. This decomposition allows each expert to independently process a subset of the input space. Notably, when all experts are combined, the overall parameter scale remains comparable to that of the original FFN layer.

\subsection{Expert Arrangement and Routing}

% 在专家排列方面，我们采用了一种多层排列专家的方法。具体来说，在将FFN层分割为更小的专家之后，我们对专家进行分层排列，每一层放置N个专家，一共放置M层。实际上，我们采取多层排列而不是单层排列专家的原因在于，单层排列专家在函数空间上是多层排列专家的一种特殊情况，证明过程详见附录A。每层放置相同数量的N个专家是为了在切割专家总量一定的情况下增加表示多样性。形式化的，第j层专家组表示为Ej, Ej={FFN_{(j-1)N + 1}, ..., FFN_{jN}}。这样实际上是对模型深度进行加深，使其可以捕捉更复杂的特征。为了方便，我们接下来使用Ej,i表示第j层专家组的第i个专家。
In terms of expert arrangement, we adopt a \textbf{multi-layer expert arrangement} strategy. Specifically, after decomposing the FFN layer into fine-grained experts, we arrange these experts in multiple sub-layers, placing $K$ experts per sub-layer across a total of $M$ sub-layers. In fact, the reason we adopt a multi-layer arrangement of experts instead of a single-layer arrangement is that the single-layer expert arrangement is a special case of the multi-layer arrangement in terms of function space. The proof of this can be found in Appendix \ref{sec:appendix A.1}. The choice to maintain a fixed number of $K$ experts per sub-layer, given a fixed total number of experts, is intended to enhance representational diversity. Formally, the expert group in sub-layer $j$ is defined as:

\begin{equation}
\mathrm{E}_{j} = \{\mathrm{FFN}_{(j-1)K + 1}, ..., \mathrm{FFN}_{jK}\}
\end{equation}
This structural design of multi-layer expert arrangement effectively increases the model’s depth, allowing it to capture more complex features. For clarity, we denote the $i$th expert in the $j$th sub-layer as:

\begin{equation}
\mathrm{E}_{j,i} = \mathrm{FFN}_{(j-1)K + i}
\end{equation}

% 在路由方式方面，我们提出了一种activation-guided routing方法。不同于MOE架构中将专家输入输入到router中，我们将专家输出输入到router中计算权重分数。这是由于我们能够获取到所有专家输出，采取这种方式可以进行更为准确的路由。在获取每个专家的权重分数后，我们并没有采取标准的softmax归一化操作，这是由于不同专家输出的竞争会加剧稀疏激活的现象。所以我们采取了sigmoid对router的权重进行非线性转换，使其在0到1的区间内，这样既能保证每个专家发挥其作用，又避免了专家之间互相影响。
Regarding the routing approach, we propose an \textbf{output-guided sigmoid routing} mechanism. Unlike the MoE architecture, where the router processes the input to determine expert selection, our method operates within a dense framework, meaning all experts are always activated. Given this, we compute weight scores based on expert outputs rather than inputs, allowing for more precise routing. Since all expert outputs are available, this approach ensures a more accurate assessment of their contributions.
Once the weight scores are obtained, we forgo the standard softmax normalization. Softmax enforces competition among experts, often amplifying sparse activation by suppressing weaker expert contributions. Instead, inspired by \citet{liu2024deepseek}, we apply a sigmoid function to nonlinearly transform the router’s weights into the range $[0,1]$. It allows each expert to contribute independently rather than being normalized in a competitive manner. This helps mitigate excessive sparsity while maintaining flexibility in expert activation.

% 需要注意的是，由于我们的方法增加了模型的深度，直接训练可能会导致梯度消失问题。为了避免这一问题，我们采用了子层残差归一化（sublayer residual normalization）操作。具体而言，为了防止训练过程中出现梯度消失并提高训练的稳定性，我们在子层与子层之间添加RMSNorm和残差连接操作。我们需要在对专家输入之前进行norm操作，在路由分数加权之后进行残差连接操作。形式上，第j子层专家组的计算过程可以表示如下式：
It is important to note that since our method increases the model's depth, direct training may lead to gradient vanishing issues. To mitigate this, we employ a \textbf{sub-layer residual normalization} operation. Specifically, to prevent gradient vanishing during training and improve training stability, we add RMSNorm and residual connection operations between sub-layers. We apply the normalization operation before the expert inputs and perform residual connections after weighting the routing scores. Formally, the computation process in the $j$th sub-layer can be expressed as follows:

% 这里，\mathrm{RMSNorm}_{j}表示第j个子层的RMSNorm模块。\bm{\hat{h}}^{l,j-1}_t表示第l层第j-1子层在t时间步的输出。同理的，最终的\bm{\hat{h}}^{l,j}_t表示第j子层专家组的输出。\tilde{h}}^{l,j}_t表示第j子层RMSNorm模块的输出。\sigma表示sigmoid激活函数。\bm{R}_{j, i}表示第j子层的路由矩阵的第i列。\bm{r}_{j,i} (\bm{\tilde{h}}^{l,j}_t)表示第j子层router对第i个专家的路由分数。

\begin{equation}
\bm{\tilde{h}}^{l,j}_t = \mathrm{RMSNorm}_{j} (\bm{\hat{h}}^{l,j-1}_t)
\end{equation}

\begin{equation}
\bm{r}_{j,i} (\bm{\tilde{h}}^{l,j}_t) = \sigma(\mathrm{E}_{j,i} (\bm{\tilde{h}}^{l,j}_t) \bm{R}_{j, i})
\end{equation}

\begin{equation}
\bm{\hat{h}}^{l,j}_t = \sum^{K}_{i=1} \bm{r}_{j,i} (\bm{\tilde{h}}^{l,j}_t) \cdot \mathrm{E}_{j,i} (\bm{\tilde{h}}^{l,j}_t)
\end{equation}

\begin{equation}
\bm{\hat{h}}^{l,j}_t :=\bm{\hat{h}}^{l,j}_t + \bm{\hat{h}}^{l,j-1}_t
\end{equation}
Here $\mathrm{RMSNorm}_{j}$ represents the RMSNorm module in the $j$th sub-layer. $\bm{\hat{h}}^{l,j-1}_t$ denotes the output of the $(j-1)$th sub-layer at time step $t$ in the $l$th layer. Similarly, the final output of the $j$th sub-layer expert group is given by $\bm{\hat{h}}^{l,j}_t$, while $\bm{\tilde{h}}^{l,j}_t$ represents the output of the RMSNorm module in the $j$th sub-layer. The function $\sigma$ denotes the sigmoid activation function. $\bm{R}_{j, i}$ refers to the 
$i$th column of the routing matrix in the $j$th sub-layer. Finally, $\bm{r}_{j,i} (\bm{\tilde{h}}^{l,j}_t)$ represents the routing score assigned by the router in the 
$j$th sub-layer to the $i$th expert.

% 第一个专家组子层将当前层多头注意力模块的输出作为输入，按照上述层内计算过程计算输出。在每个专家组子层计算完毕后，上一个子层得到的hidden states被输入到下一个子层中，直到所有专家组子层计算完毕。
Overall, the first expert group sub-layer takes the output of the current layer's MHA module as input and processes it according to the intra-layer computation described above. The hidden states produced by each expert group sub-layer are then sequentially passed to the next sub-layer until all expert group sub-layers have been processed.

\section{Experiments}
% TODO 实验部分感觉主要的问题就是baseline不够，只有一个。另外，平衡深度和宽度，这个应该是一个共识，感觉笔墨现在有点多，可以考虑倾斜到其他分析上面，这块实验稍微后置更好。后面那个分析NSAR那部分的，反而可以提前，那个跟你的motivation相呼应，其实很重要。不过当前的实验顺序这么安排也没有大问题，相当于先分析了一下深度和宽度的影响，再决定用最好的一个来分析后面的。不过确实深度和宽度这块笔墨没必要这么多。其他地方的分析反而应该更多一点。
% 我们在不同的模型尺寸和实验设置上进行了广泛的实验，并在ppl指标和下游任务上进行验证我们提出的Finedeep方法的有效性。
We conducted extensive experiments across various model sizes and configurations, evaluating perplexity results and downstream benchmarks to validate the effectiveness of our proposed Finedeep approach.

\subsection{Pre-training Dataset}

% 为了尽可能地提高训练模型的性能，我们收集了多种领域的高质量的开源预训练数据集。在通用领域，我们收集了fineweb-edu数据集，它是fineweb数据集的子集，使用教育质量分类器从fineweb数据集中筛选出大量的具有教育意义的网页数据。在数学和代码领域，我们参考olmoe收集了OpenWebMath和StarCoder数据集。OpenWebMath是从Common Crawl上中筛选和提取出来的高质量的数学文本。StarCoder包含了种类丰富的编程语言、github issue以及jupyter notebooks数据，并经过了严格的数据过滤流程。另外，合成数据已经被验证能够提高模型的性能，所以我们同时收集了cosmopedia数据集，它由合成的textbooks, blogposts, stories, posts and WikiHow articles组成，并涵盖了丰富的主题。
To maximize the performance of our trained models, we curated high-quality open-source pre-training datasets from various domains. For general-domain data, we collected the FineWeb-Edu dataset, a subset of the FineWeb dataset, which was refined using an educational quality classifier to filter and extract a large volume of high-value educational web content \cite{penedo2024fineweb}. In the domains of mathematics and code, we followed the OLMoE model to gather the OpenWebMath and StarCoder datasets \cite{muennighoff2024olmoe}. OpenWebMath consists of high-quality mathematical text filtered and extracted from Common Crawl \cite{DBLP:conf/iclr/PasterSAB24}. StarCoder includes a diverse range of programming languages, GitHub issues, and Jupyter Notebook data, undergoing a rigorous data filtering process to ensure quality \cite{DBLP:journals/tmlr/LiAZMKMMALCLZZW23}. Additionally, synthetic data has been shown to enhance model performance \cite{abdin2024phi}. To leverage this, we incorporated the Cosmopedia dataset, which consists of synthetic textbooks, blog posts, stories, posts and WikiHow articles, covering a wide range of topics \cite{benallal2024cosmopedia}.

% 我们根据附录所示对不同领域的数据集进行混合

% 在收集不同领域的预训练数据之后，我们需要对不同领域数据进行混合，参考其他开源模型技术报告和我们收集的各领域数据大小，我们最终将数据配比确定为如表1所示。由于计算资源限制，我们参照其他研究的做法，将总的预训练数据确定为100B tokens。
After collecting pre-training data from various domains, we mixed them according to the mix ratios in Appendix \ref{sec:appendix A.3}. % TODO 这里需要一句话说明一下是怎么得到ratio的，而不是直接说茶这个配比，这样不严谨，reviewer可能问你配比哪里来的。
Our mixing strategy was informed by technical reports from other open-source models, as well as the dataset sizes we gathered across different domains. Given computational resource constraints, we set the total pre-training data size to 100B tokens, following best practices from related studies \cite{DBLP:conf/acl/DaiDZXGCLZYWXLH24, su2024cartesianmoe, DBLP:conf/nips/Xie0DDLLLL0Y23}.

% 在进行预训练实验前，我们还需要对数据进行分词处理。具体的，我们使用llama3的词表大小为128k的tokenizer对混合的数据进行分词处理，限制最大长度为1024。
Before conducting pre-training experiments, we also preprocessed the data for tokenization. Specifically, we utilized LLaMA 3's tokenizer, which has a vocabulary size of 128K, to tokenize the mixed dataset while enforcing a maximum sequence length of 1,024 \cite{dubey2024llama}.

\subsection{Experimental Setup}

% 跟随xx等人的研究，我们设置了两种尺寸大小的模型进行预训练，分别是small、medium和large设置。
% 对于small模型设置，总参数量为665M，hidden size为1024，原始intermediate size为4096，注意力头数为16，总层数设置为24。
% 对于medium模型设置，总参数量为1.6B，hidden size为2048，原始intermediate size为8192，注意力头数为8，总层数设置为16。
% 对于large模型设置，总参数量为7.5B，hidden size为4096，原始intermediate size为11008，注意力头数为32，总层数设置为32。
% 对于所有尺寸设置的模型，我们采取FusedAdam优化器，并将学习率设置为3e-4，weight decay设置为1e-1，初始化模型权重正态分布的标准差设置为1e-2，rmsnorm层的epsilon设置为1e-5。
Following the studies by \citet{DBLP:conf/icml/BidermanSABOHKP23} and \citet{su2024mile}, we conducted pre-training experiments with three model configurations: \emph{Small}, \emph{Medium} and \emph{Large}. The \emph{Small} model setup consists of 665M parameters, the \emph{Medium} model setup has 1.6B parameters, and the \emph{Large} model setup includes 7.5B parameters. Specific training configurations are detailed in the Appendix \ref{sec:appendix A.5}.
% Small模型设置包含665M模型参数，medium模型设置包含1.6B模型参数，large模型设置包含7.5B模型参数

% \begin{itemize}
% \item \textbf{Small}: This model configuration contains 665M parameters, with a hidden size of 1,024, a raw intermediate size of 4,096, 16 attention heads, and 24 layers in total.
% \item \textbf{Medium}: This configuration consists of 1.6B parameters, featuring a hidden size of 2,048, a raw intermediate size of 8,192, 8 attention heads, and 16 layers.
% \item \textbf{Large}: This setting comprises 7.5B parameters, with a hidden size of 4,096, a raw intermediate size of 11,008, 32 attention heads, and 32 layers.
% \end{itemize}

% For all model sizes, we employed the FusedAdam optimizer with a learning rate of 3e-4. We set weight decay to 1e-1. Model weights were initialized from a normal distribution with a standard deviation of 1e-2, and the RMSNorm epsilon was set to 1e-5.

% 对于两种不同的模型尺寸设置，我们分别对其进行切割不同子层数量和每个子层不同专家数量的实验来验证我们方法的有效性。特别的，在small模型设置下，我们控制专家排列子层数为2层不变，分别进行每层排列4/8/16个专家的实验。类似的，在medium模型设置下，我们也进行同样设置的实验，不同的是，我们另外控制总切割专家数为16不变，分别进行专家排列子层数为4/8层的设置。在large模型设置下，我们使用总切割专家数为16，专家排列子层数为2的实验设置来证明提出方法的有效性。
To evaluate the effectiveness of our method, we conducted experiments with varying numbers of sub-layers and experts per sub-layer. Specifically, in the \textit{Small} model setup, we fixed the number of expert sub-layers to 2 and conducted experiments with 4, 8, and 16 experts per sub-layer. The same experimental configuration was applied to the \textit{Medium} model. Additionally, we performed experiments where the total number of experts was fixed at 16, while varying the number of expert sub-layers to 4 and 8, respectively. For the \textit{Large} model, we maintained the total number of experts at 16 and set the number of expert sub-layers to 2, demonstrating the effectiveness of our proposed method. % TODO expert arrangement sub-layer这个词太长了，很难理解，还不如直接改成expert sublayer或者直接就是sublayer，可以double check一下，全文复查一下

% 在评测方面，我们同时进行PPL评测和benchmark评测两种评测。在PPL评测方面，我们跟随xx等人的研究，测试模型在包含22个domain的pile测试集上的困惑度指标。在benchmark评测方面，我们使用lm-evaluation-harness库进行评测。我们同时进行了判别式任务和生成式任务的评测，对于判别式任务，我们报告0-shot的结果，对于生成式任务，我们报告5-shot的结果。我们评测的benchmark覆盖了多个领域，能够测试模型各个方面的能力，包括reading comprehension的SQuAD V2，测试language understading的lambada，测试commonsense reasoning的arc challenge/hellaswag/piqa/siqa/winogrande，以及测试closed-book question answering的Natural Question和TriviaQA。
For the evaluation, we conducted both PPL and benchmark evaluations. For the PPL evaluation, we followed the approach outlined by \citet{DBLP:conf/acl/DaiDZXGCLZYWXLH24}, testing the model's perplexity on the pile test set. In terms of benchmark evaluations, we used the lm-evaluation-harness \cite{eval-harness} tool library for our evaluation. We performed both discriminative and generative tasks, reporting zero-shot results for the discriminative tasks and five-shot results for the generative tasks. The benchmarks we collected cover a broad range of domains to assess various aspects of the model's capabilities. A detailed description can be found in Appendix \ref{sec:appendix A.4}.

% 具体介绍详见附录A.4 

\begin{table}[]
\centering
\resizebox{0.5\textwidth}{!}{%
\begin{tabular}{lccccc}
\toprule
\textbf{Model}             & \textbf{\begin{tabular}[c]{@{}c@{}}Sub-layer \\ Counts\end{tabular}} & \textbf{\begin{tabular}[c]{@{}c@{}}Experts per \\ Sub-layer\end{tabular}} & \textbf{Pile PPL} (↓) \\ \midrule
\multicolumn{4}{l}{\textit{Small}}                                                                        \\ \midrule
Standard Dense        & N/A                        & N/A                              & 14.36             \\
Finedeep (Ours) & M=2                        & K=4                              & 14.28             \\
                      & M=2                        & K=8                              & \textbf{14.16}    \\
                      & M=2                        & K=16                             & 14.18             \\ \midrule
\multicolumn{4}{l}{\textit{Medium}}                                                                       \\ \midrule
Standard Dense        & N/A                        & N/A                              & 12.42             \\
Finedeep (Ours) & M=2                        & K=4                              & 12.24             \\
                      & M=2                        & K=8                              & 12.23             \\
                      & M=2                        & K=16                             & 12.24             \\
                      & M=4                        & K=4                              & \textbf{12.13}    \\
                      & M=8                        & K=2                              & 12.17             \\ \midrule
\multicolumn{4}{l}{\textit{Large}}                                                                        \\ \midrule
Standard Dense        & N/A                        & N/A                              & 10.15             \\
Finedeep (Ours) & M=2                        & K=8                              & \textbf{10.08}    \\ \bottomrule
\end{tabular}
}
\caption{Perplexity results for models with different configurations. The best results are highlighted in \textbf{bold}. $M$ denotes the number of sub-layers in the expert arrangement, $K$ represents the number of experts per sub-layer.}
\label{tab:table2}
\end{table}
% PPL实验结果。最好的结果被加粗表示。M表示专家排列的子层数量。N表示每个子层排列的专家数量。

\begin{table*}[]
\centering
\resizebox{1.0\textwidth}{!}{%
\begin{tabular}{lccccccccccc}
\toprule
\textbf{Model}               & \textbf{SQuAD} & \textbf{LAMBADA} & \textbf{ARC} & \textbf{HellaSwag} & \textbf{PIQA} & \textbf{SIQA} & \textbf{Wino} & \textbf{NaturalQs} & \textbf{TriviaQA} & \textbf{AVG} \\ 
\midrule
\multicolumn{11}{l}{\textit{Small}} \\ 
\midrule
Standard Dense               & 6.22  & 41.14  & 33.87  & 49.79  & 70.78  & 41.15  & 54.14  & 7.04  & 20.79  & 36.10  \\ 
Finedeep \textcolor{lightgray}{M=2/K=4}    & {\ul 7.34}  & \textbf{42.23}  & {\ul 34.13}  & 50.44  & 70.51  & 40.23  & 55.01  & {\ul 6.79}  & \textbf{21.79}  & 36.50  \\ 
Finedeep \textcolor{lightgray}{M=2/K=8}    & \textbf{9.53}  & {\ul 42.01}  & \textbf{36.09}  & {\ul 50.58}  & {\ul 71.22}  & {\ul 40.79}  & \textbf{56.27}  & 6.51  & {\ul 21.54}  & \textbf{37.17}  \\ 
Finedeep \textcolor{lightgray}{M=2/K=16}   & 6.89  & 41.76  & 33.79  & \textbf{50.60}  & \textbf{71.87}  & \textbf{41.20}  & {\ul 55.64}  & \textbf{7.04}  & 21.25  & {\ul 36.67}  \\ 
\midrule
\multicolumn{11}{l}{\textit{Medium}} \\ 
\midrule
Standard Dense               & 7.16  & 46.52  & 38.99  & 56.45  & 73.67  & 42.43  & 56.91  & 8.56  & 28.25  & 39.88  \\ 
Finedeep \textcolor{lightgray}{M=2/K=4}    & \textbf{15.65}  & 46.59  & 39.68  & {\ul 57.52}  & 72.96  & 41.50  & 56.35  & {\ul 9.28}  & 29.43  & 41.00  \\ 
Finedeep \textcolor{lightgray}{M=2/K=8}    & {\ul 14.95}  & \textbf{48.30}  & {\ul 39.93}  & 57.49  & \textbf{74.10}  & \textbf{43.60}  & 56.83  & 8.53  & 28.99  & \textbf{41.41}  \\ 
Finedeep \textcolor{lightgray}{M=2/K=16}   & 14.76  & 47.99  & 39.68  & 57.24  & 73.45  & 42.17  & 56.99  & 8.81  & 29.39  & 41.16  \\ 
Finedeep \textcolor{lightgray}{M=4/K=4}    & 12.22  & 47.80  & \textbf{40.19}  & \textbf{58.11}  & 73.72  & {\ul 42.48}  & {\ul 59.19}  & 8.23  & \textbf{30.20}  & {\ul 41.35}  \\ 
Finedeep \textcolor{lightgray}{M=8/K=2}    & 12.64  & {\ul 48.19}  & 38.91  & 57.29  & {\ul 73.99}  & 41.97  & \textbf{59.27}  & \textbf{9.78}  & {\ul 29.98}  & 41.34  \\ 
\midrule
\multicolumn{11}{l}{\textit{Large}} \\ 
\midrule
Standard Dense               & 19.50  & 55.00  & \textbf{46.93}  & 66.05  & 76.28  & 43.50  & 62.19  & 13.74  & 42.26  & 47.27  \\ 
Finedeep \textcolor{lightgray}{M=2/K=8}    & \textbf{19.92}  & \textbf{56.26}  & 45.90  & \textbf{66.25}  & \textbf{76.99}  & \textbf{43.86}  & \textbf{62.43}  & \textbf{14.27}  & \textbf{43.33}  & \textbf{47.69}  \\ 
\bottomrule
\end{tabular}
}
\caption{Benchmark results for models with different configurations. The best results are highlighted in \textbf{bold}, while the second-best results are {\ul underlined}. Here, $M$ denotes the number of sub-layers in the expert arrangement, $K$ represents the number of experts per sub-layer. The AVG metric represents the average of the different benchmark results.}
\label{tab:table3}
\end{table*}

% 不同设置的模型在bencnmark上的结果。最好的结果被加粗表示。第二好的结果被下划线表示。M表示专家排列的子层数量。N表示每个子层排列的专家数量。

% \subsection{Main Results}

% % 在PPL指标和下游benchmark任务上的结果显示，我们的方法显著优于传统的稠密模型。另外我们还探究了专家排列子层数和每层排列的专家数两个因素对我们的方法的影响。
% Results on PPL metrics and downstream benchmark tasks demonstrate that our method significantly outperforms traditional dense models. Additionally, we investigate the impact of two key factors: the number of expert arrangement sub-layers and the number of experts per sub-layer.

\subsection{Perplexity Results}

\begin{table}[]
\centering
\resizebox{0.5\textwidth}{!}{%
\begin{tabular}{lccc}
\toprule
                  & \multicolumn{1}{c}{\begin{tabular}[c]{@{}c@{}}w/o multi experts\\ \textcolor{lightgray}{M=2/K=1}\end{tabular}} & \multicolumn{1}{c}{\begin{tabular}[c]{@{}c@{}}w/o multi sub-layers\\ \textcolor{lightgray}{M=1/K=16}\end{tabular}} & \multicolumn{1}{c}{\begin{tabular}[c]{@{}c@{}}Finedeep\\ \textcolor{lightgray}{M=2/K=8}\end{tabular}} \\ \midrule
PPL (↓)              & 12.42                        & 12.42                        & \textbf{12.23}              \\ \midrule
SQuAD          & 10.11                        & 12.65                        & \textbf{14.95}              \\
LAMBADA           & 46.48                        & 45.06                        & \textbf{48.30}              \\
ARC     & 39.08                        & 38.65                        & \textbf{39.93}              \\
HellaSwag         & 56.58                        & 56.46                        & \textbf{57.49}              \\
PIQA              & 73.50                        & 72.96                        & \textbf{74.10}              \\
SIQA              & 41.45                        & 40.74                        & \textbf{43.60}              \\
Wino        & 57.30                        & \textbf{57.38}               & 56.83                       \\
NaturalQs & \textbf{9.11}                & 8.39                         & 8.53                        \\
TriviaQA          & 28.61                        & 28.61                        & \textbf{28.99}              \\
AVG               & 40.25                        & 40.10                        & \textbf{41.41}              \\ \bottomrule
\end{tabular}
}
\caption{Ablation experiment results on the impact of multiple experts per sub-layer and multiple sub-layers.}
\label{tab:table4}
\end{table}

% 在不同模型尺度上，我们提出的方法在pile测试集上的PPL都显著优于传统的稠密模型，如图2所示。
% 特别的，对于每层排列专家数的最优选择，我们发现，在small模型设置下，在确保专家排列子层数量不变的情况下，每层排列8个专家的设置比每层排列4/16个专家的设置效果更好，medium模型设置也表现出了相似的结果。这说明了适当增加每层排列专家数量能够提升模型表现，但是过度增多切分专家数量则会损害模型性能。
% 对于专家排列子层数量的最优选择，我们发现，在medium设置下，将专家排列子层数由2增大到4，但是不改变切分专家总量，可以提升模型性能，但是进一步的，将专家排列子层数由4增大到8，模型性能则有所下降。这说明了适当增加专家排列子层数对模型表现有利，但是过度增加模型排列子层数则对模型表现有害。
% 总的来说，我们追求专家排列过程中宽度与深度的平衡。
Our proposed method significantly outperforms traditional dense models in terms of PPL on the PILE test set across different model scales, as shown in Table \ref{tab:table2}. Specifically, for the optimal choice of the number of experts per sub-layer, we find that in the \emph{Small} model setup, when the number of expert sub-layers is kept constant, the configuration with 8 experts per sub-layer outperforms the configurations with 4 or 16 experts per sub-layer. A similar trend was observed in the \emph{Medium} model setup. This suggests that appropriately increasing the number of experts per sub-layer can enhance model performance, as a sufficient number of experts allows the sub-layer to capture more complex features. However, excessive increases in the number of experts can reintroduce the sparse activation problem, leading to inefficient activation utilization and diminished performance. % TODO 后面跟一句为什么是这样，可以解释说，expert多了之后，还是会有稀疏激活的问题，少了的话，却不够在这一层学习充分，类似的。

Regarding the optimal choice of the number of expert sub-layers, we find that in the \emph{Medium} setup, increasing the number of expert sub-layers from 2 to 4, while keeping the total number of experts constant, enhances model performance. However, further increasing the number of sub-layers from 4 to 8 results in a performance drop. This indicates that while increasing the number of expert sub-layers can benefit model performance, excessive depth in the model can be detrimental. The underlying reason is analogous to the earlier observation, as keeping a fixed total number of experts, too many sub-layers will result in fewer experts per sub-layer, and too few sub-layers will concentrate too many experts in each sub-layer. % TODO 同上面一句话解释为什么，其实不外乎就是同等参数下，层数多了，每一层的expert数量变少，导致效果不好，层数少了，则每层expert比较多。本质上原因跟上一段是一样的。

In summary, we aim to strike a balance between width and depth in the expert arrangement process.

\subsection{Benchmark Results}

% 在不同大小的模型设置下，我们的方法在覆盖多领域的benchmark上的表现都优于传统的稠密模型，如表3所示。我们发现，不同领域benchmark的AVG指标表现出和ppl指标相似的趋势，基本上是专家排列子层数为2，每层排列专家数为8的设置或者是专家排列子层数为4，每层排列专家数为4的设置表现最优。这进一步验证了我们的方法在宽度和深度达到平衡时取得最优结果的结论。
As shown in Table \ref{tab:table3}, our method outperforms the traditional dense model across a range of benchmarks covering multiple domains. We observe that the AVG metrics for these benchmarks follow a similar trend to the PPL metrics. Specifically, configurations with 2 expert sub-layers and 8 experts per layer, or 4 expert sub-layers and 4 experts per layer, yield the best performance. This reinforces the conclusion that our method achieves optimal results when there is a balanced trade-off between width and depth.

\section{Analysis}

\subsection{Ablation Study}

% 为了进一步证明每个子层切分多个专家和排列多个子层的组合的必要性，我们在medium的模型尺寸下进行了消融实验，实验结果如表4所示。
% 首先，我们验证每个子层排列多个专家的必要性。具体来说，我们将子层数设置为2，每个子层的专家数设置为1。特别的，由于这里每层只有一个专家，所以我们去掉了为每个专家分配权重的router，实验结果表示，这样的设置在PPL和benchmark上的结果远远不如我们的方法，甚至在某些benchmark上的结果不如baseline的效果，这说明每个子层排列多个专家的必要性。
% 进一步的，我们也验证了设置多个专家子层的必要性。具体来说，我们将子层数设置为1，这个子层的专家数设置为16。结果表明，采取这样的实验设置取得的效果也不理想。这说明设置多个专家子层的必要性。
% 另外，这两组消融实验同样进一步验证了每个子层排列的专家数和设置的子层数达到平衡时取得最优结果的结论，因为这两组消融实验的设置都是极端不平衡的样例，所以取得了较差效果。

To further demonstrate the necessity of splitting multiple experts per sub-layer and arranging multiple sub-layers, we conducted ablation experiments using the \emph{Medium} size model. Experimental results are presented in Table \ref{tab:table4}.

First, we validated the necessity of arranging multiple experts within each sub-layer. Specifically, we set the number of sub-layers to 2 and assigned only one expert per sub-layer. Notably, since there was only one expert per layer in this setup, we removed the router responsible for assigning weights to each expert. The results show that this configuration performs significantly worse than our method in terms of both PPL and benchmark evaluations. In some benchmarks, its performance is even inferior to the baseline, highlighting the importance of arranging multiple experts within each sub-layer.

Furthermore, we verified the necessity of using multiple expert sub-layers. In this experiment, we set the number of sub-layers to 1 while assigning 16 experts within that single sub-layer. The results indicate that this setup also leads to suboptimal performance, further emphasizing the importance of arranging multiple sub-layers.

Additionally, these two ablation studies reinforce our conclusion that achieving a balance between the number of experts per sub-layer and the number of sub-layers leads to optimal results. Both experimental configurations represent extreme imbalances, which result in poor performance.

% 首先，我们将子层数设置为2，每个子层的专家设置为1，特别的，由于这里每层只有一个专家，所以我们去掉了router和norm层。实验结果表明，这样的设置效果远远不如我们的方法，并且甚至在某些方面不如baseline的效果，这证明了每个子层切分多个专家的必要性。进一步的，我们将专家排列子层数设置为1，并且切分16个专家，类似的，这样的实验设置取得的效果也并不理想，这说明排列多个子层的必要性。
% To further illustrate the necessity of incorporating multiple experts per sub-layer and arranging multiple sub-layer combinations, we conducted ablation experiments, with results presented in Table 4. First, we configured the model with two sub-layers, each containing only one expert. Notably, since there was only one expert per sub-layer, we removed the router and normalization layers. The experimental results indicate that this setup performed significantly worse than our approach and, in some cases, even underperformed compared to the baseline model. This finding highlights the importance of utilizing multiple experts per sub-layer. Furthermore, we conducted an experiment where we set the number of expert arrangement sub-layers to one while increasing the number of experts per sub-layer to 16. This configuration also yielded suboptimal results, reinforcing the necessity of arranging multiple sub-layers in our method.

\subsection{Routing Scores Computation: Sigmoid vs. Softmax Comparison}

% 我们提出的方法在计算路由分数时，对router的输出经过sigmoid函数处理后作为最终的路由分数，但一般来说，在传统的moe结构中都是对路由分数进行softmax归一化处理，如下式所示：
Our proposed method computes the final routing score by applying a sigmoid function to the router's output, whereas traditional MoE structures typically use softmax normalization \cite{fedus2022switch}, as shown in the following equation:

\begin{equation}
\bm{r}_{j,i} (\bm{\tilde{h}}^{l,j}_t) = \frac{\mathrm{exp}(\mathrm{E}_{j,i} (\bm{\tilde{h}}^{l,j}_t) \bm{R}_{j, i})}{\sum^N_{i=1}\mathrm{exp}(\mathrm{E}_{j,i} (\bm{\tilde{h}}^{l,j}_t) \bm{R}_{j, i})}
\end{equation}

% 我们对比了两种计算最终路由分数的方法，实验结果如表5所示。PPL和benchmark结果表明，Finedeep所采用的sigmoid函数效果更好。
We compared these two approaches for computing routing scores using the \emph{Medium} model, and experimental results presented in Table \ref{tab:table5} demonstrate that the sigmoid-based routing in Finedeep achieves superior performance in terms of both PPL and benchmark results. This improvement can be attributed to the fact that softmax enforces competition among experts, whereas sigmoid allows each expert to contribute independently. As a result, the sigmoid method reduces unnecessary competition, leading to a more balanced utilization of model capacity. Given that our approach activates all expert parameters, maintaining this balance is particularly crucial for maximizing performance.

\begin{table}[]
\centering
\resizebox{0.3\textwidth}{!}{%
\begin{tabular}{lcc}
\toprule
                  & \begin{tabular}[c]{@{}c@{}}softmax\\ \textcolor{lightgray}{M=2/K=8}\end{tabular} & \begin{tabular}[c]{@{}c@{}}sigmoid\\ \textcolor{lightgray}{M=2/K=8}\end{tabular} \\ \midrule
PPL (↓)               & 12.27                                                      & \textbf{12.23}                                            \\ \midrule
SQuAD          & \textbf{14.99}                                             & 14.95                                                     \\
LAMBADA           & 47.04                                                      & \textbf{48.30}                                            \\
ARC     & 39.85                                                      & \textbf{39.93}                                            \\
HellaSwag         & 56.80                                                      & \textbf{57.49}                                            \\
PIQA              & 72.63                                                      & \textbf{74.10}                                            \\
SIQA              & 42.37                                                      & \textbf{43.60}                                            \\
Wino        & \textbf{57.30}                                             & 56.83                                                     \\
NaturalQs & \textbf{8.67}                                              & 8.53                                                      \\
TriviaQA          & 27.94                                                      & \textbf{28.99}                                            \\
AVG               & 40.84                                                      & \textbf{41.41}                                            \\ \bottomrule
\end{tabular}
}
\caption{Experimental results comparing the Softmax and Sigmoid methods for computing routing scores.}
\label{tab:table5}
\end{table}

\subsection{Params and FLOPs of Finedeep}

% 我们计算了我们提出的方法的参数量和浮点运算数，并与传统dense架构进行对比，如表6结果所示。我们的方法虽然额外加入了router模块和rmsnorm模块，但是这些模块的参数量相比模型总参数量很小，计算表明，我们的模型在不同尺寸的模型上大概只比传统的dense模型增大了0.03% - 0.06% 的参数量和0.03% - 0.08% 的浮点运算数，这几乎是可以忽略不计的。我们的方法在参数量和浮点运算数和传统dense模型基本持平的情况下，达到了更优的效果，这进一步说明我们的方法的有效性。
We compute the number of parameters and floating-point operations (FLOPs) of our proposed method and compare them with those of the traditional dense architecture, as shown in Table \ref{tab:table6}. Although our method introduces additional components, such as the router module and RMSNorm, the parameter overhead from these modules is minimal relative to the total model size. Our calculations indicate that across different model scales, our approach increases the parameter count by only 0.03\%–0.06\% and FLOPs by 0.03\%–0.08\% compared to the traditional dense model, which is an almost negligible difference. Despite maintaining nearly the same parameter count and FLOPs as the dense baseline, our method achieves significantly better performance, further demonstrating its effectiveness.

\begin{table}[]
\centering
\resizebox{0.4\textwidth}{!}{%
\begin{tabular}{lcc}
\toprule
                       & \textbf{Params}      & \textbf{GFLOPs}      \\ \midrule
\textit{Small}         & \multicolumn{1}{l}{} & \multicolumn{1}{l}{} \\ \midrule
standard dense         & 665.37 M             & 138.33               \\
Finedeep \textcolor{lightgray}{M=2/K=8} & 665.79 M             & 138.44               \\ \midrule
\textit{Medium}        & \multicolumn{1}{l}{} & \multicolumn{1}{l}{} \\ \midrule
standard dense         & 1.5992 B             & 344.29               \\
Finedeep \textcolor{lightgray}{M=2/K=8} & 1.5997 B             & 344.37               \\
Finedeep \textcolor{lightgray}{M=4/K=4} & 1.6002 B             & 344.51               \\ \midrule
\textit{Large}         &                      &                      \\ \midrule
standard dense         & 7.5269 B             & 1801.00                 \\
Finedeep \textcolor{lightgray}{M=2/K=8} & 7.5292 B             & 1801.46              \\ \bottomrule
\end{tabular}
}
\caption{Comparison of parameter count and GFLOPs between our method and standard dense architectures. GFLOPs (Giga floating-point operations) are calculated by processing input samples with a batch size of 1 and a sequence length of 128.}
\label{tab:table6}
\end{table}

\subsection{Mitigating Sparse Activation with Finedeep}

\begin{figure}[t]
\centering
\centerline{\includegraphics[scale=0.28]{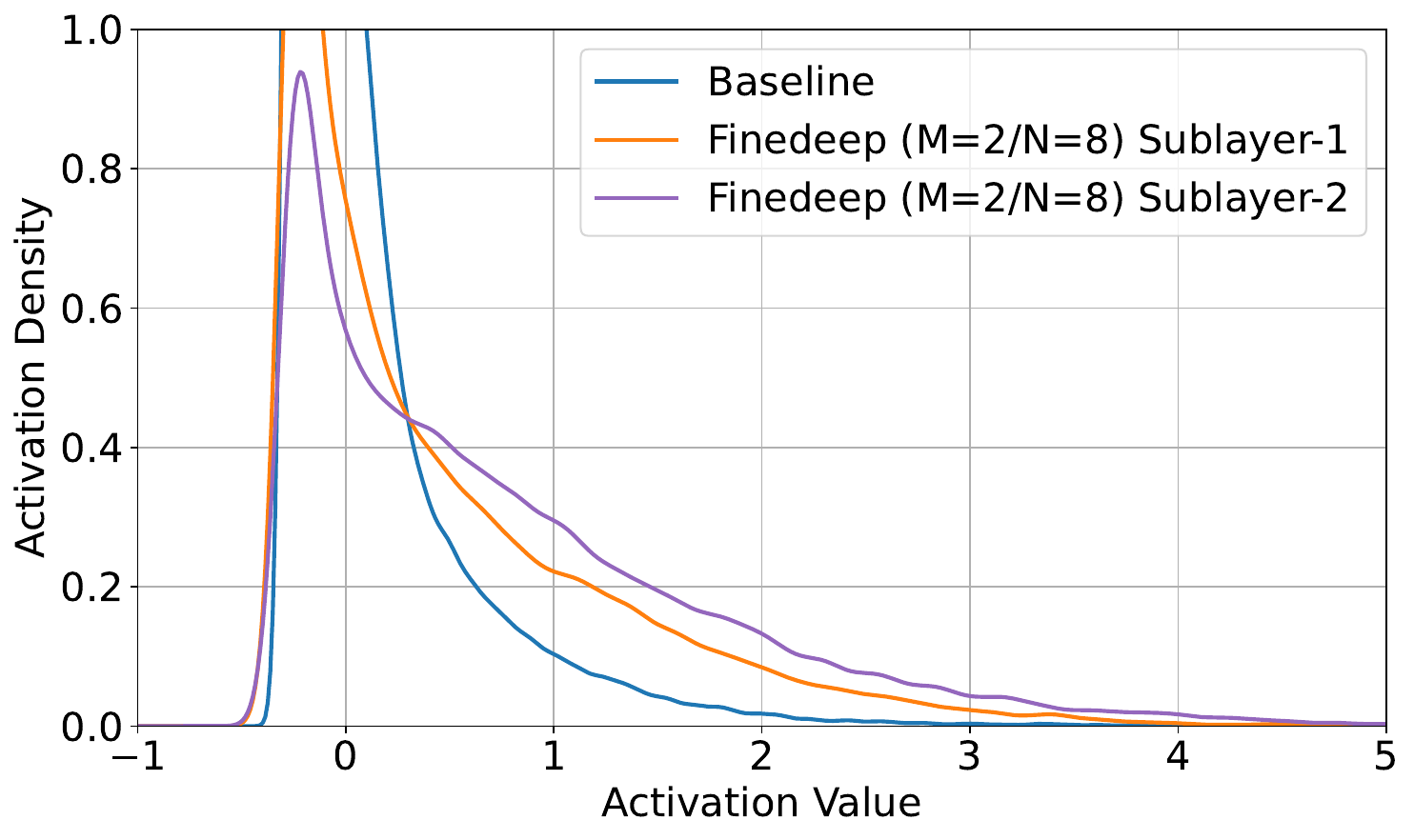}}
\caption{Output distributions of the activation functions for Finedeep and the baseline model.}
\label{fig3}
\end{figure}

% 我们实证性地发现，Finedeep确实能够缓解稀疏激活的现象，如图3所示。具体的，我们采取了专家排列子层数为2，每个子层排列8个专家的设置，并可视化第一个子层和第二个子层中的激活函数的输出的分布，与传统dense模型的激活函数的输出分布进行对比。我们发现，Finedeep激活函数的输出分布相比baseline更加均匀，0附近的值更少，较大的值分布的更多。值得注意的是，随着层数的加深，这种均匀的现象更加明显。
% 为了更直观的说明我们的方法缓解了稀疏激活问题，我们定义了一种非稀疏激活率的指标，它的计算方式如下公式所示：
We empirically observe that Finedeep effectively mitigates the issue of sparse activation, as illustrated in Figure \ref{fig3}. Specifically, we adopt a configuration with 2 expert sub-layers, each containing 8 experts, and visualize the distribution of the activation function outputs in the first and second sub-layers. This is compared to the activation function output distributions of the traditional dense model. Our findings indicate that the output distribution of Finedeep is more homogeneous, with fewer values concentrated around 0 and a broader distribution of larger values. Notably, this uniformity becomes more pronounced as the model depth increases.

\begin{figure}[t]
\centering
\centerline{\includegraphics[scale=0.31]{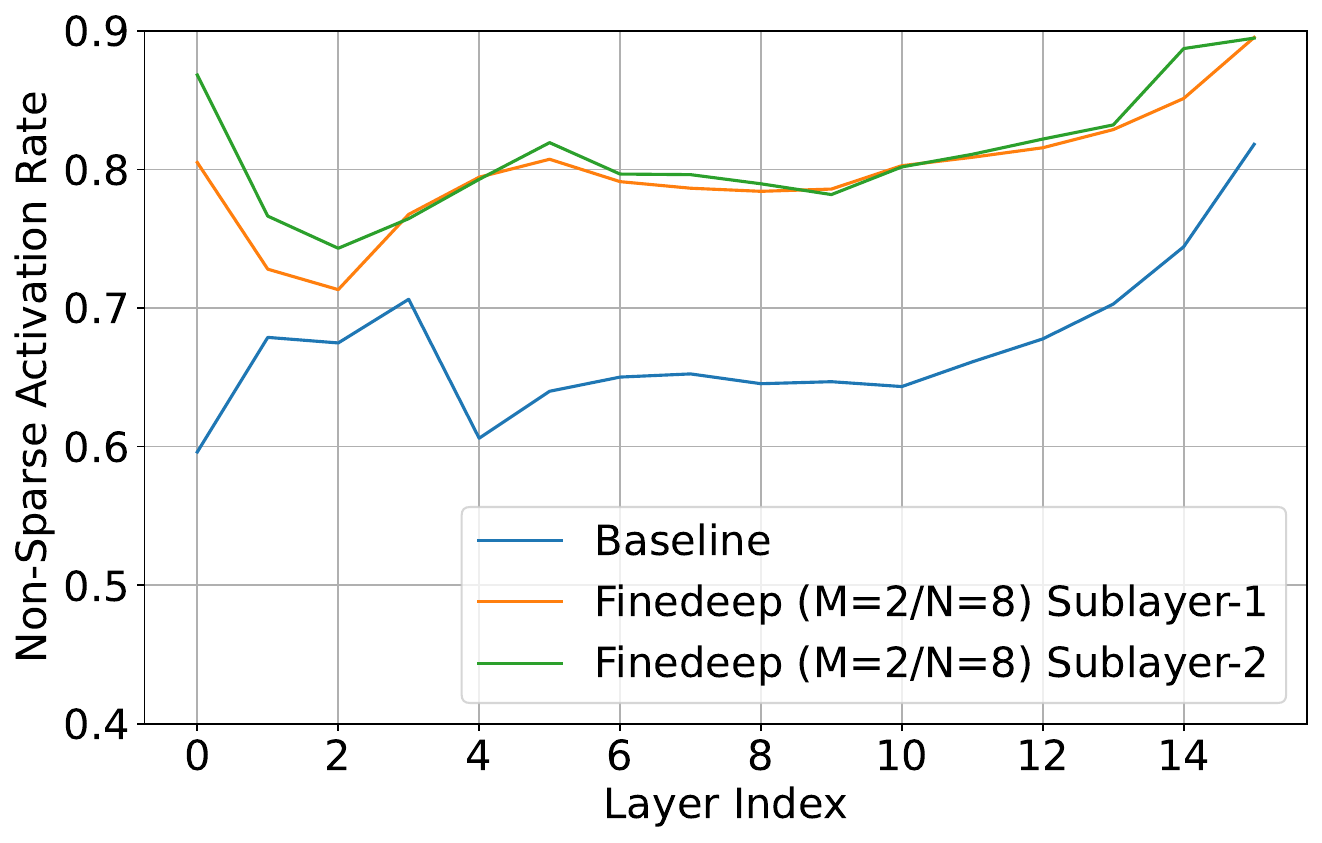}}
\caption{Variation of $\mathrm{NSAR}_{0.1}$ metrics across different model layers.}
\label{fig4}
\end{figure}

To better illustrate that our approach mitigates the sparse activation problem, we introduce a metric called NSAR (i.e., \textbf{N}on-\textbf{S}parse \textbf{A}ctivation \textbf{R}ate), defined as follows:

$$
\mathrm{N S A R}_{\tau}=\frac{\sum_{i, j} \mathbb{I}\left(\left|\textbf{A}_{i, j}\right|>\tau\right)}{B \times H}
$$

\begin{equation}
\text{where} \quad \mathbb{I}\left(\left|\textbf{A}_{i, j}\right|>\tau\right)=\left\{\begin{array}{ll}
1, & \text { if}\left|\textbf{A}_{i, j}\right|>\tau \\
0, & \text { else }
\end{array}\right.
\end{equation}
% 其中A为模型某一层的激活矩阵，B为批量大小，H为神经元数量, \tau为预设的阈值。我们将模型不同层的NSAR可视化为图4，显然的，我们的方法在不同层上都缓解了baseline模型的稀疏激活现象。
% 通过缓解稀疏激活的问题，我们的方法对激活值有更高的利用率，从而进一步拓展激活值的表示空间，提升模型的表示能力，如附录B所示。
Here, $\textbf{A}$ represents the activation matrix of a model layer, $B$ is the batch size, $H$ denotes the number of neurons, and $\tau$ is a predefined threshold. In Figure \ref{fig4}, we visualize the $\mathrm{NSAR}_{0.1}$ metric across different model layers, clearly demonstrating that our method effectively mitigates the sparse activation phenomenon in the traditional dense model. By alleviating the sparse activation problem, our method increases the utilization of activation values, thereby expanding their representation capacity and enhancing the model's ability to represent complex features, as demonstrated in Appendix \ref{sec:appendix A.2}.

\section{Conclusion}

% 为了缓解现有的dense模型中出现的稀疏激活现象，我们提出了一种新的dense架构Finedeep。它通过将现有的dense架构的FFN层分割成多个专家，并通过多子层排列专家的方式拓宽模型的深度，在子层内使用router来决定每个专家贡献程度。我们在多个模型尺寸上进行广泛实验，PPL和benchmark结果表明，我们的方法显著优于现有的dense架构。另外，我们发现当专家排列子层数和每个子层的专家数达到平衡时效果最优。进一步的，我们通过消融实验证明排列多个子层和子层内排列多个专家的必要性，并实证性地证明了我们提出的方法确实能够缓解稀疏激活现象。
To address the sparse activation phenomenon observed in existing dense models, we have presented a novel architecture called Finedeep. It enhances the model's depth by splitting the FFN layer of traditional dense architectures into multiple experts, arranged across sub-layers. Routers within these sub-layers are employed to control the contribution of each expert. We conduct extensive experiments across multiple model sizes, and the PPL and benchmark results demonstrate that our method significantly outperforms existing dense architectures with identical parameter counts. Additionally, we find that the model performs optimally when the number of expert sub-layers and the number of experts per sub-layer are balanced. Through ablation experiments, we further highlight the importance of both arranging multiple sub-layers and distributing multiple experts within each sub-layer. Our empirical results show that Finedeep effectively mitigates the issue of sparse activation.

\section*{Limitations}
% 由于计算资源限制，我们仅对所有的模型设置训练了100B的tokens，未探究训练更多tokens的情况，另外，我们最大的模型尺寸设置为7.5B，未探究更大模型的情况。另外，我们认为dense模型中的稀疏激活的现象还有进一步缓解的空间。我们将这些留给后人研究。
Due to computational resource constraints, we trained all model configurations on only 100B tokens and did not explore the impact of training on a larger token budget. Additionally, our largest model size was limited to 7.5B parameters, leaving the potential benefits of scaling to larger models unexplored. Furthermore, while our approach mitigates sparse activation in dense models, we believe there is still room for further improvement. We leave this for our future research.

\bibliography{custom}

\begin{thebibliography}{39}
\providecommand{\natexlab}[1]{#1}

\bibitem[{Abdin et~al.(2024)Abdin, Aneja, Behl, Bubeck, Eldan, Gunasekar, Harrison, Hewett, Javaheripi, Kauffmann et~al.}]{abdin2024phi}
Marah Abdin, Jyoti Aneja, Harkirat Behl, S{\'e}bastien Bubeck, Ronen Eldan, Suriya Gunasekar, Michael Harrison, Russell~J Hewett, Mojan Javaheripi, Piero Kauffmann, et~al. 2024.
\newblock Phi-4 technical report.
\newblock \emph{arXiv preprint arXiv:2412.08905}.

\bibitem[{Achiam et~al.(2023)Achiam, Adler, Agarwal, Ahmad, Akkaya, Aleman, Almeida, Altenschmidt, Altman, Anadkat et~al.}]{achiam2023gpt}
Josh Achiam, Steven Adler, Sandhini Agarwal, Lama Ahmad, Ilge Akkaya, Florencia~Leoni Aleman, Diogo Almeida, Janko Altenschmidt, Sam Altman, Shyamal Anadkat, et~al. 2023.
\newblock Gpt-4 technical report.
\newblock \emph{arXiv preprint arXiv:2303.08774}.

\bibitem[{Ben~Allal et~al.(2024)Ben~Allal, Lozhkov, Penedo, Wolf, and von Werra}]{benallal2024cosmopedia}
Loubna Ben~Allal, Anton Lozhkov, Guilherme Penedo, Thomas Wolf, and Leandro von Werra. 2024.
\newblock \href {https://huggingface.co/datasets/HuggingFaceTB/cosmopedia} {Cosmopedia}.

\bibitem[{Biderman et~al.(2023)Biderman, Schoelkopf, Anthony, Bradley, O'Brien, Hallahan, Khan, Purohit, Prashanth, Raff, Skowron, Sutawika, and van~der Wal}]{DBLP:conf/icml/BidermanSABOHKP23}
Stella Biderman, Hailey Schoelkopf, Quentin~Gregory Anthony, Herbie Bradley, Kyle O'Brien, Eric Hallahan, Mohammad~Aflah Khan, Shivanshu Purohit, USVSN~Sai Prashanth, Edward Raff, Aviya Skowron, Lintang Sutawika, and Oskar van~der Wal. 2023.
\newblock \href {https://proceedings.mlr.press/v202/biderman23a.html} {Pythia: {A} suite for analyzing large language models across training and scaling}.
\newblock In \emph{International Conference on Machine Learning, {ICML} 2023, 23-29 July 2023, Honolulu, Hawaii, {USA}}, volume 202 of \emph{Proceedings of Machine Learning Research}, pages 2397--2430. {PMLR}.

\bibitem[{Bisk et~al.(2020)Bisk, Zellers, Gao, Choi et~al.}]{bisk2020piqa}
Yonatan Bisk, Rowan Zellers, Jianfeng Gao, Yejin Choi, et~al. 2020.
\newblock Piqa: Reasoning about physical commonsense in natural language.
\newblock In \emph{Proceedings of the AAAI Conference on Artificial Intelligence}, volume~34, pages 7432--7439.

\bibitem[{Brown et~al.(2020)Brown, Mann, Ryder, Subbiah, Kaplan, Dhariwal, Neelakantan, Shyam, Sastry, Askell, Agarwal, Herbert{-}Voss, Krueger, Henighan, Child, Ramesh, Ziegler, Wu, Winter, Hesse, Chen, Sigler, Litwin, Gray, Chess, Clark, Berner, McCandlish, Radford, Sutskever, and Amodei}]{DBLP:conf/nips/BrownMRSKDNSSAA20}
Tom~B. Brown, Benjamin Mann, Nick Ryder, Melanie Subbiah, Jared Kaplan, Prafulla Dhariwal, Arvind Neelakantan, Pranav Shyam, Girish Sastry, Amanda Askell, Sandhini Agarwal, Ariel Herbert{-}Voss, Gretchen Krueger, Tom Henighan, Rewon Child, Aditya Ramesh, Daniel~M. Ziegler, Jeffrey Wu, Clemens Winter, Christopher Hesse, Mark Chen, Eric Sigler, Mateusz Litwin, Scott Gray, Benjamin Chess, Jack Clark, Christopher Berner, Sam McCandlish, Alec Radford, Ilya Sutskever, and Dario Amodei. 2020.
\newblock \href {https://proceedings.neurips.cc/paper/2020/hash/1457c0d6bfcb4967418bfb8ac142f64a-Abstract.html} {Language models are few-shot learners}.
\newblock In \emph{Advances in Neural Information Processing Systems 33: Annual Conference on Neural Information Processing Systems 2020, NeurIPS 2020, December 6-12, 2020, virtual}.

\bibitem[{Clark et~al.(2018)Clark, Cowhey, Etzioni, Khot, Sabharwal, Schoenick, and Tafjord}]{DBLP:journals/corr/abs-1803-05457}
Peter Clark, Isaac Cowhey, Oren Etzioni, Tushar Khot, Ashish Sabharwal, Carissa Schoenick, and Oyvind Tafjord. 2018.
\newblock \href {https://arxiv.org/abs/1803.05457} {Think you have solved question answering? try arc, the {AI2} reasoning challenge}.
\newblock \emph{CoRR}, abs/1803.05457.

\bibitem[{Dai et~al.(2024)Dai, Deng, Zhao, Xu, Gao, Chen, Li, Zeng, Yu, Wu, Xie, Li, Huang, Luo, Ruan, Sui, and Liang}]{DBLP:conf/acl/DaiDZXGCLZYWXLH24}
Damai Dai, Chengqi Deng, Chenggang Zhao, R.~X. Xu, Huazuo Gao, Deli Chen, Jiashi Li, Wangding Zeng, Xingkai Yu, Y.~Wu, Zhenda Xie, Y.~K. Li, Panpan Huang, Fuli Luo, Chong Ruan, Zhifang Sui, and Wenfeng Liang. 2024.
\newblock \href {https://doi.org/10.18653/V1/2024.ACL-LONG.70} {Deepseekmoe: Towards ultimate expert specialization in mixture-of-experts language models}.
\newblock In \emph{Proceedings of the 62nd Annual Meeting of the Association for Computational Linguistics (Volume 1: Long Papers), {ACL} 2024, Bangkok, Thailand, August 11-16, 2024}, pages 1280--1297. Association for Computational Linguistics.

\bibitem[{Dai et~al.(2022)Dai, Dong, Hao, Sui, Chang, and Wei}]{dai2022knowledge}
Damai Dai, Li~Dong, Yaru Hao, Zhifang Sui, Baobao Chang, and Furu Wei. 2022.
\newblock Knowledge neurons in pretrained transformers.
\newblock In \emph{Proceedings of the 60th Annual Meeting of the Association for Computational Linguistics (Volume 1: Long Papers)}, pages 8493--8502.

\bibitem[{Dubey et~al.(2024)Dubey, Jauhri, Pandey, Kadian, Al-Dahle, Letman, Mathur, Schelten, Yang, Fan et~al.}]{dubey2024llama}
Abhimanyu Dubey, Abhinav Jauhri, Abhinav Pandey, Abhishek Kadian, Ahmad Al-Dahle, Aiesha Letman, Akhil Mathur, Alan Schelten, Amy Yang, Angela Fan, et~al. 2024.
\newblock The llama 3 herd of models.
\newblock \emph{arXiv preprint arXiv:2407.21783}.

\bibitem[{Fedus et~al.(2022)Fedus, Zoph, and Shazeer}]{fedus2022switch}
William Fedus, Barret Zoph, and Noam Shazeer. 2022.
\newblock Switch transformers: Scaling to trillion parameter models with simple and efficient sparsity.
\newblock \emph{Journal of Machine Learning Research}, 23(120):1--39.

\bibitem[{Gao et~al.(2024)Gao, Tow, Abbasi, Biderman, Black, DiPofi, Foster, Golding, Hsu, Le~Noac'h, Li, McDonell, Muennighoff, Ociepa, Phang, Reynolds, Schoelkopf, Skowron, Sutawika, Tang, Thite, Wang, Wang, and Zou}]{eval-harness}
Leo Gao, Jonathan Tow, Baber Abbasi, Stella Biderman, Sid Black, Anthony DiPofi, Charles Foster, Laurence Golding, Jeffrey Hsu, Alain Le~Noac'h, Haonan Li, Kyle McDonell, Niklas Muennighoff, Chris Ociepa, Jason Phang, Laria Reynolds, Hailey Schoelkopf, Aviya Skowron, Lintang Sutawika, Eric Tang, Anish Thite, Ben Wang, Kevin Wang, and Andy Zou. 2024.
\newblock \href {https://doi.org/10.5281/zenodo.12608602} {A framework for few-shot language model evaluation}.

\bibitem[{Geva et~al.(2021)Geva, Schuster, Berant, and Levy}]{geva2021transformer}
Mor Geva, Roei Schuster, Jonathan Berant, and Omer Levy. 2021.
\newblock Transformer feed-forward layers are key-value memories.
\newblock In \emph{Proceedings of the 2021 Conference on Empirical Methods in Natural Language Processing}, pages 5484--5495.

\bibitem[{Hoffmann et~al.(2022)Hoffmann, Borgeaud, Mensch, Buchatskaya, Cai, Rutherford, de~Las~Casas, Hendricks, Welbl, Clark, Hennigan, Noland, Millican, van~den Driessche, Damoc, Guy, Osindero, Simonyan, Elsen, Rae, Vinyals, and Sifre}]{DBLP:journals/corr/abs-2203-15556}
Jordan Hoffmann, Sebastian Borgeaud, Arthur Mensch, Elena Buchatskaya, Trevor Cai, Eliza Rutherford, Diego de~Las~Casas, Lisa~Anne Hendricks, Johannes Welbl, Aidan Clark, Tom Hennigan, Eric Noland, Katie Millican, George van~den Driessche, Bogdan Damoc, Aurelia Guy, Simon Osindero, Karen Simonyan, Erich Elsen, Jack~W. Rae, Oriol Vinyals, and Laurent Sifre. 2022.
\newblock \href {https://doi.org/10.48550/ARXIV.2203.15556} {Training compute-optimal large language models}.
\newblock \emph{CoRR}, abs/2203.15556.

\bibitem[{Joshi et~al.(2017)Joshi, Choi, Weld, and Zettlemoyer}]{joshi2017triviaqa}
Mandar Joshi, Eunsol Choi, Daniel~S Weld, and Luke Zettlemoyer. 2017.
\newblock Triviaqa: A large scale distantly supervised challenge dataset for reading comprehension.
\newblock In \emph{Proceedings of the 55th Annual Meeting of the Association for Computational Linguistics (Volume 1: Long Papers)}, pages 1601--1611.

\bibitem[{Kaplan et~al.(2020)Kaplan, McCandlish, Henighan, Brown, Chess, Child, Gray, Radford, Wu, and Amodei}]{DBLP:journals/corr/abs-2001-08361}
Jared Kaplan, Sam McCandlish, Tom Henighan, Tom~B. Brown, Benjamin Chess, Rewon Child, Scott Gray, Alec Radford, Jeffrey Wu, and Dario Amodei. 2020.
\newblock \href {https://arxiv.org/abs/2001.08361} {Scaling laws for neural language models}.
\newblock \emph{CoRR}, abs/2001.08361.

\bibitem[{Kwiatkowski et~al.(2019)Kwiatkowski, Palomaki, Redfield, Collins, Parikh, Alberti, Epstein, Polosukhin, Devlin, Lee et~al.}]{kwiatkowski2019natural}
Tom Kwiatkowski, Jennimaria Palomaki, Olivia Redfield, Michael Collins, Ankur Parikh, Chris Alberti, Danielle Epstein, Illia Polosukhin, Jacob Devlin, Kenton Lee, et~al. 2019.
\newblock Natural questions: A benchmark for question answering research.
\newblock \emph{Transactions of the Association for Computational Linguistics}, 7:452--466.

\bibitem[{Lepikhin et~al.(2021)Lepikhin, Lee, Xu, Chen, Firat, Huang, Krikun, Shazeer, and Chen}]{DBLP:conf/iclr/LepikhinLXCFHKS21}
Dmitry Lepikhin, HyoukJoong Lee, Yuanzhong Xu, Dehao Chen, Orhan Firat, Yanping Huang, Maxim Krikun, Noam Shazeer, and Zhifeng Chen. 2021.
\newblock \href {https://openreview.net/forum?id=qrwe7XHTmYb} {Gshard: Scaling giant models with conditional computation and automatic sharding}.
\newblock In \emph{9th International Conference on Learning Representations, {ICLR} 2021, Virtual Event, Austria, May 3-7, 2021}. OpenReview.net.

\bibitem[{Li et~al.(2023)Li, Allal, Zi, Muennighoff, Kocetkov, Mou, Marone, Akiki, Li, Chim, Liu, Zheltonozhskii, Zhuo, Wang, Dehaene, Davaadorj, Lamy{-}Poirier, Monteiro, Shliazhko, Gontier, Meade, Zebaze, Yee, Umapathi, Zhu, Lipkin, Oblokulov, Wang, V, Stillerman, Patel, Abulkhanov, Zocca, Dey, Zhang, Fahmy, Bhattacharyya, Yu, Singh, Luccioni, Villegas, Kunakov, Zhdanov, Romero, Lee, Timor, Ding, Schlesinger, Schoelkopf, Ebert, Dao, Mishra, Gu, Robinson, Anderson, Dolan{-}Gavitt, Contractor, Reddy, Fried, Bahdanau, Jernite, Ferrandis, Hughes, Wolf, Guha, von Werra, and de~Vries}]{DBLP:journals/tmlr/LiAZMKMMALCLZZW23}
Raymond Li, Loubna~Ben Allal, Yangtian Zi, Niklas Muennighoff, Denis Kocetkov, Chenghao Mou, Marc Marone, Christopher Akiki, Jia Li, Jenny Chim, Qian Liu, Evgenii Zheltonozhskii, Terry~Yue Zhuo, Thomas Wang, Olivier Dehaene, Mishig Davaadorj, Joel Lamy{-}Poirier, Jo{\~{a}}o Monteiro, Oleh Shliazhko, Nicolas Gontier, Nicholas Meade, Armel Zebaze, Ming{-}Ho Yee, Logesh~Kumar Umapathi, Jian Zhu, Benjamin Lipkin, Muhtasham Oblokulov, Zhiruo Wang, Rudra~Murthy V, Jason~T. Stillerman, Siva~Sankalp Patel, Dmitry Abulkhanov, Marco Zocca, Manan Dey, Zhihan Zhang, Nour Fahmy, Urvashi Bhattacharyya, Wenhao Yu, Swayam Singh, Sasha Luccioni, Paulo Villegas, Maxim Kunakov, Fedor Zhdanov, Manuel Romero, Tony Lee, Nadav Timor, Jennifer Ding, Claire Schlesinger, Hailey Schoelkopf, Jan Ebert, Tri Dao, Mayank Mishra, Alex Gu, Jennifer Robinson, Carolyn~Jane Anderson, Brendan Dolan{-}Gavitt, Danish Contractor, Siva Reddy, Daniel Fried, Dzmitry Bahdanau, Yacine Jernite, Carlos~Mu{\~{n}}oz Ferrandis, Sean Hughes, Thomas Wolf,
  Arjun Guha, Leandro von Werra, and Harm de~Vries. 2023.
\newblock \href {https://openreview.net/forum?id=KoFOg41haE} {Starcoder: may the source be with you!}
\newblock \emph{Trans. Mach. Learn. Res.}, 2023.

\bibitem[{Li et~al.()Li, You, Bhojanapalli, Li, Rawat, Reddi, Ye, Chern, Yu, Guo et~al.}]{lilazy}
Zonglin Li, Chong You, Srinadh Bhojanapalli, Daliang Li, Ankit~Singh Rawat, Sashank~J Reddi, Ke~Ye, Felix Chern, Felix Yu, Ruiqi Guo, et~al.
\newblock The lazy neuron phenomenon: On emergence of activation sparsity in transformers.
\newblock In \emph{The Eleventh International Conference on Learning Representations}.

\bibitem[{Liu et~al.(2024)Liu, Feng, Xue, Wang, Wu, Lu, Zhao, Deng, Zhang, Ruan et~al.}]{liu2024deepseek}
Aixin Liu, Bei Feng, Bing Xue, Bingxuan Wang, Bochao Wu, Chengda Lu, Chenggang Zhao, Chengqi Deng, Chenyu Zhang, Chong Ruan, et~al. 2024.
\newblock Deepseek-v3 technical report.
\newblock \emph{arXiv preprint arXiv:2412.19437}.

\bibitem[{Luo et~al.(2024)Luo, Song, Han, Chen, Xiao, Liu, and Sun}]{luo2024sparsing}
Yuqi Luo, Chenyang Song, Xu~Han, Yingfa Chen, Chaojun Xiao, Zhiyuan Liu, and Maosong Sun. 2024.
\newblock Sparsing law: Towards large language models with greater activation sparsity.
\newblock \emph{arXiv preprint arXiv:2411.02335}.

\bibitem[{Muennighoff et~al.(2024)Muennighoff, Soldaini, Groeneveld, Lo, Morrison, Min, Shi, Walsh, Tafjord, Lambert et~al.}]{muennighoff2024olmoe}
Niklas Muennighoff, Luca Soldaini, Dirk Groeneveld, Kyle Lo, Jacob Morrison, Sewon Min, Weijia Shi, Pete Walsh, Oyvind Tafjord, Nathan Lambert, et~al. 2024.
\newblock Olmoe: Open mixture-of-experts language models.
\newblock \emph{CoRR}.

\bibitem[{Paperno et~al.(2016)Paperno, Kruszewski, Lazaridou, Pham, Bernardi, Pezzelle, Baroni, Boleda, and Fern{\'a}ndez}]{paperno2016lambada}
Denis Paperno, Germ{\'a}n Kruszewski, Angeliki Lazaridou, Ngoc-Quan Pham, Raffaella Bernardi, Sandro Pezzelle, Marco Baroni, Gemma Boleda, and Raquel Fern{\'a}ndez. 2016.
\newblock The lambada dataset: Word prediction requiring a broad discourse context.
\newblock In \emph{Proceedings of the 54th Annual Meeting of the Association for Computational Linguistics (Volume 1: Long Papers)}, pages 1525--1534.

\bibitem[{Paster et~al.(2024)Paster, Santos, Azerbayev, and Ba}]{DBLP:conf/iclr/PasterSAB24}
Keiran Paster, Marco~Dos Santos, Zhangir Azerbayev, and Jimmy Ba. 2024.
\newblock \href {https://openreview.net/forum?id=jKHmjlpViu} {Openwebmath: An open dataset of high-quality mathematical web text}.
\newblock In \emph{The Twelfth International Conference on Learning Representations, {ICLR} 2024, Vienna, Austria, May 7-11, 2024}. OpenReview.net.

\bibitem[{Penedo et~al.(2024)Penedo, Kydl{\'\i}{\v{c}}ek, Lozhkov, Mitchell, Raffel, Von~Werra, Wolf et~al.}]{penedo2024fineweb}
Guilherme Penedo, Hynek Kydl{\'\i}{\v{c}}ek, Anton Lozhkov, Margaret Mitchell, Colin Raffel, Leandro Von~Werra, Thomas Wolf, et~al. 2024.
\newblock The fineweb datasets: Decanting the web for the finest text data at scale.
\newblock \emph{arXiv preprint arXiv:2406.17557}.

\bibitem[{Rajpurkar et~al.(2018)Rajpurkar, Jia, and Liang}]{rajpurkar2018know}
Pranav Rajpurkar, Robin Jia, and Percy Liang. 2018.
\newblock Know what you don’t know: Unanswerable questions for squad.
\newblock In \emph{Proceedings of the 56th Annual Meeting of the Association for Computational Linguistics (Volume 2: Short Papers)}, pages 784--789.

\bibitem[{Sakaguchi et~al.(2020)Sakaguchi, Le~Bras, Bhagavatula, and Choi}]{sakaguchi2020winogrande}
Keisuke Sakaguchi, Ronan Le~Bras, Chandra Bhagavatula, and Yejin Choi. 2020.
\newblock Winogrande: An adversarial winograd schema challenge at scale.
\newblock In \emph{Proceedings of the AAAI Conference on Artificial Intelligence}, volume~34, pages 8732--8740.

\bibitem[{Sap et~al.(2019)Sap, Rashkin, Chen, Le~Bras, and Choi}]{sap2019social}
Maarten Sap, Hannah Rashkin, Derek Chen, Ronan Le~Bras, and Yejin Choi. 2019.
\newblock Social iqa: Commonsense reasoning about social interactions.
\newblock In \emph{Proceedings of the 2019 Conference on Empirical Methods in Natural Language Processing and the 9th International Joint Conference on Natural Language Processing (EMNLP-IJCNLP)}, pages 4463--4473.

\bibitem[{Su et~al.(2024{\natexlab{a}})Su, Lin, Baixue, Chen, Hu, Zhou, Ding, and Xing}]{su2024mile}
Zhenpeng Su, Zijia Lin, Baixue Baixue, Hui Chen, Songlin Hu, Wei Zhou, Guiguang Ding, and W~Xing. 2024{\natexlab{a}}.
\newblock Mile loss: a new loss for mitigating the bias of learning difficulties in generative language models.
\newblock In \emph{Findings of the Association for Computational Linguistics: NAACL 2024}, pages 250--262.

\bibitem[{Su et~al.(2024{\natexlab{b}})Su, Wu, Lin, Xiong, Lv, Ma, Chen, Hu, and Ding}]{su2024cartesianmoe}
Zhenpeng Su, Xing Wu, Zijia Lin, Yizhe Xiong, Minxuan Lv, Guangyuan Ma, Hui Chen, Songlin Hu, and Guiguang Ding. 2024{\natexlab{b}}.
\newblock Cartesianmoe: Boosting knowledge sharing among experts via cartesian product routing in mixture-of-experts.
\newblock \emph{arXiv preprint arXiv:2410.16077}.

\bibitem[{Touvron et~al.(2023{\natexlab{a}})Touvron, Lavril, Izacard, Martinet, Lachaux, Lacroix, Rozi{\`{e}}re, Goyal, Hambro, Azhar, Rodriguez, Joulin, Grave, and Lample}]{DBLP:journals/corr/abs-2302-13971}
Hugo Touvron, Thibaut Lavril, Gautier Izacard, Xavier Martinet, Marie{-}Anne Lachaux, Timoth{\'{e}}e Lacroix, Baptiste Rozi{\`{e}}re, Naman Goyal, Eric Hambro, Faisal Azhar, Aur{\'{e}}lien Rodriguez, Armand Joulin, Edouard Grave, and Guillaume Lample. 2023{\natexlab{a}}.
\newblock \href {https://doi.org/10.48550/ARXIV.2302.13971} {Llama: Open and efficient foundation language models}.
\newblock \emph{CoRR}, abs/2302.13971.

\bibitem[{Touvron et~al.(2023{\natexlab{b}})Touvron, Martin, Stone, Albert, Almahairi, Babaei, Bashlykov, Batra, Bhargava, Bhosale et~al.}]{touvron2023llama}
Hugo Touvron, Louis Martin, Kevin Stone, Peter Albert, Amjad Almahairi, Yasmine Babaei, Nikolay Bashlykov, Soumya Batra, Prajjwal Bhargava, Shruti Bhosale, et~al. 2023{\natexlab{b}}.
\newblock Llama 2: Open foundation and fine-tuned chat models.
\newblock \emph{arXiv preprint arXiv:2307.09288}.

\bibitem[{Xie et~al.(2023)Xie, Pham, Dong, Du, Liu, Lu, Liang, Le, Ma, and Yu}]{DBLP:conf/nips/Xie0DDLLLL0Y23}
Sang~Michael Xie, Hieu Pham, Xuanyi Dong, Nan Du, Hanxiao Liu, Yifeng Lu, Percy Liang, Quoc~V. Le, Tengyu Ma, and Adams~Wei Yu. 2023.
\newblock \href {http://papers.nips.cc/paper\_files/paper/2023/hash/dcba6be91359358c2355cd920da3fcbd-Abstract-Conference.html} {Doremi: Optimizing data mixtures speeds up language model pretraining}.
\newblock In \emph{Advances in Neural Information Processing Systems 36: Annual Conference on Neural Information Processing Systems 2023, NeurIPS 2023, New Orleans, LA, USA, December 10 - 16, 2023}.

\bibitem[{Yang et~al.(2024{\natexlab{a}})Yang, Yang, Zhang, Hui, Zheng, Yu, Li, Liu, Huang, Wei et~al.}]{yang2024qwen2}
An~Yang, Baosong Yang, Beichen Zhang, Binyuan Hui, Bo~Zheng, Bowen Yu, Chengyuan Li, Dayiheng Liu, Fei Huang, Haoran Wei, et~al. 2024{\natexlab{a}}.
\newblock Qwen2. 5 technical report.
\newblock \emph{arXiv preprint arXiv:2412.15115}.

\bibitem[{Yang et~al.(2024{\natexlab{b}})Yang, Qi, Gu, Wang, Gao, and Xu}]{yang2024xmoe}
Yuanhang Yang, Shiyi Qi, Wenchao Gu, Chaozheng Wang, Cuiyun Gao, and Zenglin Xu. 2024{\natexlab{b}}.
\newblock Xmoe: Sparse models with fine-grained and adaptive expert selection.
\newblock In \emph{Findings of the Association for Computational Linguistics ACL 2024}, pages 11664--11674.

\bibitem[{Zellers et~al.(2019)Zellers, Holtzman, Bisk, Farhadi, and Choi}]{zellers2019hellaswag}
Rowan Zellers, Ari Holtzman, Yonatan Bisk, Ali Farhadi, and Yejin Choi. 2019.
\newblock Hellaswag: Can a machine really finish your sentence?
\newblock In \emph{Proceedings of the 57th Annual Meeting of the Association for Computational Linguistics}, pages 4791--4800.

\bibitem[{Zhang et~al.(2024)Zhang, Zeng, Wang, and Lu}]{zhang2024tinyllama}
Peiyuan Zhang, Guangtao Zeng, Tianduo Wang, and Wei Lu. 2024.
\newblock Tinyllama: An open-source small language model.
\newblock \emph{arXiv preprint arXiv:2401.02385}.

\bibitem[{Zhang et~al.(2022)Zhang, Lin, Liu, Li, Sun, and Zhou}]{zhang2022moefication}
Zhengyan Zhang, Yankai Lin, Zhiyuan Liu, Peng Li, Maosong Sun, and Jie Zhou. 2022.
\newblock Moefication: Transformer feed-forward layers are mixtures of experts.
\newblock In \emph{Findings of the Association for Computational Linguistics: ACL 2022}, pages 877--890.

\end{thebibliography}

\clearpage

\appendix

\section{Appendix}

\subsection{Relationship between multi-layer and single-layer experts arrangements}
\label{sec:appendix A.1}

% 我们的目标是证明单层排列专家是多层排列专家的一种特殊情况，即多层排列专家的表示的函数空间包含单层排列专家的表示的函数空间。为了方便，我们假设子层数为2，其他子层数的情况也可以类比。
Our goal is to demonstrate that single-layer expert arrangement is a special case of multi-layer expert arrangement. In other words, the function space represented by multi-layer expert arrangement encompasses that of the single-layer approach. For simplicity, we consider the case where the number of sub-layers is 2, though the same reasoning can be extended to other configurations.

% 我们可以将第二层子层的输出$\bm{\hat{h}}^{l,j}_t$表示如下:
We can express the output of the second sub-layer $\bm{\hat{h}}^{l,2}_t$ as follows:

\begin{equation}
\bm{\hat{h}}^{l,2}_t = \sum^{K}_{i=1} \bm{r}_{2,i} (\bm{\hat{h}}^{l,1}_t) \cdot \mathrm{E}_{2,i} (\bm{\hat{h}}^{l,1}_t) + \bm{\hat{h}}^{l,1}_t
\label{eq10}
\end{equation}
% 这里为了方便，我们直接将第一个子层的输出$\bm{\hat{h}}^{l,1}_t$作为专家的输入。公式的最后一项$\bm{\hat{h}}^{l,1}_t$是作为第二个子层的残差加入的。
Here, for convenience, we directly use the output of the first sub-layer, $\bm{\hat{h}}^{l,1}_t$, as the input to the expert. The last term of the formula $\bm{\hat{h}}^{l,1}_t$ is added as the residual of the second sub-layer.

% 实际上，$\bm{\hat{h}}^{l,1}_t$也可以进一步展开为第一个子层的残差$\bm{\hat{h}}^{l,0}_t$和第一个子层专家加权求和的结果$\bm{\hat{h}}^{l,1}_t$，所以可以表示为如下形式：
In fact, $\bm{\hat{h}}^{l,1}_t$ can also be further expanded into the residuals of the first sub-layer $\bm{\hat{h}}^{l,0}_t$ and the result of the weighted summation of the experts of the first sub-layer $\bm{\hat{h}}^{l,1}_t$. So Equation \ref{eq10} can be expressed in the following form:

\begin{equation}
\bm{\hat{h}}^{l,2}_t = \sum^{K}_{i=1} \bm{r}_{2,i} \cdot \mathrm{E}_{2,i} (\bm{\hat{h}}^{l,1}_t + \bm{\hat{h}}^{l,0}_t) + \bm{\hat{h}}^{l,1}_t + \bm{\hat{h}}^{l,0}_t
\label{eq11}
\end{equation}

% 我们进一步对专家的计算过程$\mathrm{E}_{2,i}$进行展开，表示为如下形式：
We further expand the expert's computational procedure $\mathrm{E}_{2,i}$, expressed in the following form:

\begin{equation}
\begin{aligned}
\bm{\hat{h}}^{l,2}_t = \sum^{K}_{i=1} \bm{r}_{2,i} \cdot (\mathrm{E}_{2,i} (\bm{\hat{h}}^{l,1}_t) + \mathrm{E}_{2,i} (\bm{\hat{h}}^{l,0}_t) + \Delta_1) \\ + \bm{\hat{h}}^{l,1}_t + \bm{\hat{h}}^{l,0}_t
\end{aligned}
\label{eq12}
\end{equation}
% 由于在专家的前向传播过程中存在激活函数，所以这里我们使用$\Delta_1$代表非线性函数影响的补偿，从另一种角度也可以理解为是对eq11的一阶泰勒展开。
Since there is an activation function in the forward propagation process of the expert, here we use $\Delta_1$ to represent the compensation for the effect of the nonlinear function. Equation \ref{eq12} can also be interpreted in another way as a first-order Taylor expansion of equation \ref{eq11}.

% 我们进一步对eq12进行展开，如下式所示：
We further expand Equation \ref{eq12} as shown in the following equation:

\begin{equation}
\begin{aligned}
\bm{\hat{h}}^{l,2}_t = \sum^{K}_{i=1} \bm{r}_{2,i} \cdot \mathrm{E}_{2,i} (\bm{\hat{h}}^{l,1}_t) + \sum^{K}_{i=1} \bm{r}_{2,i} \cdot \mathrm{E}_{2,i} (\bm{\hat{h}}^{l,0}_t) \\ + \sum^{K}_{i=1} \bm{r}_{2,i} \cdot \Delta_1 + \bm{\hat{h}}^{l,1}_t + \bm{\hat{h}}^{l,0}_t
\end{aligned}
\label{eq13}
\end{equation}

% $\bm{\hat{h}}^{l,1}_t)$可以表示为第一个子层专家计算的过程，所以上式可以变换为：
$\bm{\hat{h}}^{l,1}_t$ can be expressed as the process of the first sub-layer expert computation, so the above equation can be transformed as:

\begin{small}
\begin{equation}
\begin{aligned}
\bm{\hat{h}}^{l,2}_t & = A + \sum^{K}_{i=1} \bm{r}_{2,i} \cdot \mathrm{E}_{2,i} (\bm{\hat{h}}^{l,0}_t) + \sum^{K}_{i=1} \bm{r}_{1,i} \cdot \mathrm{E}_{1,i} (\bm{\hat{h}}^{l,0}_t) + \bm{\hat{h}}^{l,0}_t \\
& \text{where} \quad A = \sum^{K}_{i=1} \bm{r}_{2,i} \cdot \mathrm{E}_{2,i} (\sum^{K}_{i=1} \bm{r}_{1,i} \cdot \mathrm{E}_{1,i} (\bm{\hat{h}}^{l,0}_t)) \\ & \quad \quad \quad \quad + \sum^{K}_{i=1} \bm{r}_{2,i} \cdot \Delta_1 + \bm{\hat{h}}^{l,1}_t
\end{aligned}
\label{eq14}
\end{equation}
\end{small}

% 我们可以把上式中除A项以外的所有项看做单层排列所有专家的计算过程。综上所述，我们成功证明单层排列专家是多层排列专家的一种特殊情况，所以我们采取多层排列专家的方法。
We can regard all the terms in the above equation except term $A$ as the calculation process of single-layer expert arrangement. To summarize, we successfully show that single-layer expert arrangement is a special case of multi-layer expert arrangement, so we take the method of multi-layer expert arrangement.

\subsection{Activation Clustering}
\label{sec:appendix A.2}

\begin{figure*}[t]
\centering
\centerline{\includegraphics[scale=0.3]{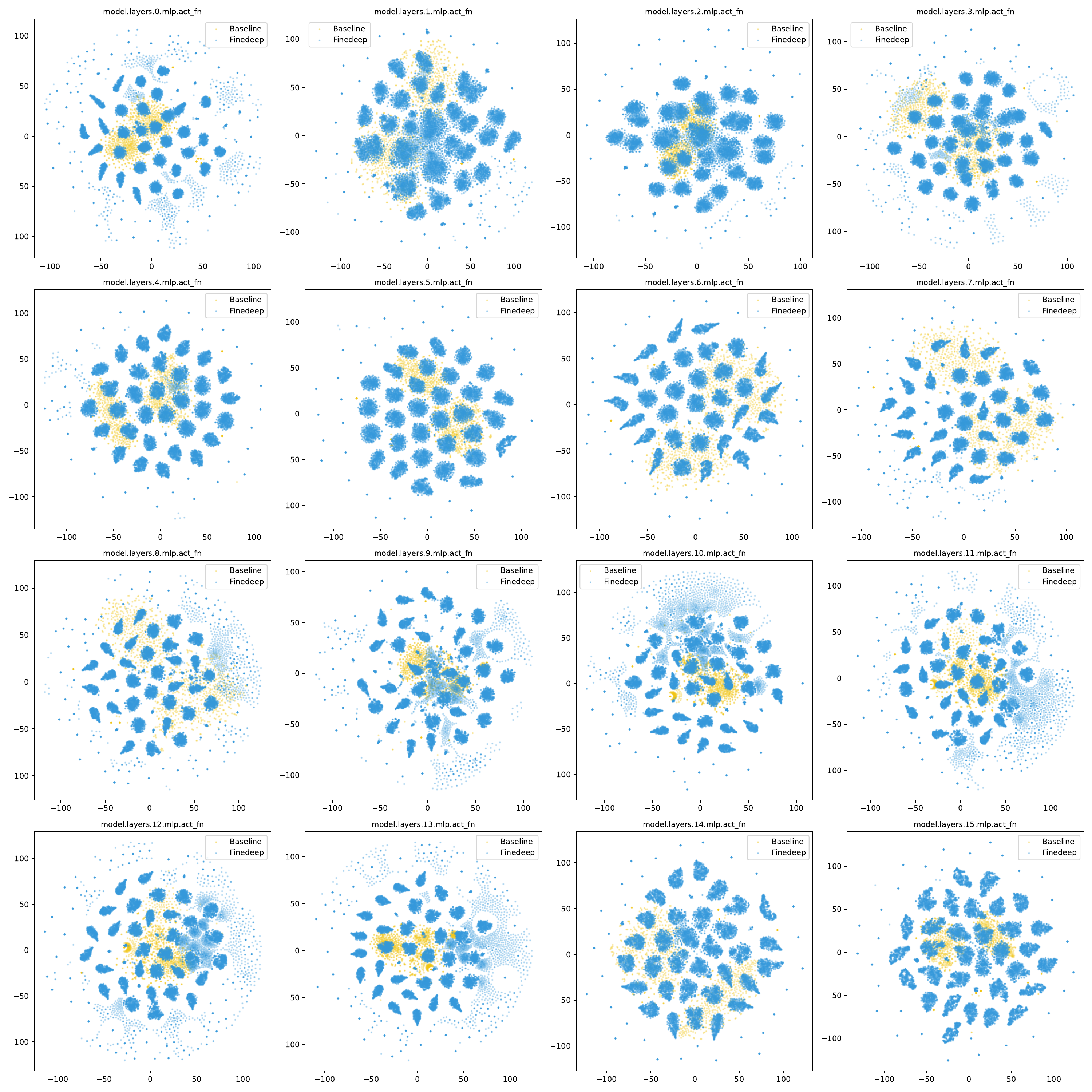}}
\caption{T-SNE clustering of activation values from different layers in the traditional dense model and the model trained with the Finedeep method.}
\label{fig5}
\end{figure*}

% 我们的方法通过缓解稀疏激活现象，提高了激活值的利用率，从而进一步拓展了激活值的表示空间。这也是我们的方法为什么比传统dense模型表现更好的原因。为了证明这一点，我们对传统dense模型和我们的方法训练出的模型的激活值进行t-sne降维，如图5所示。
Our method enhances the utilization of activation values by addressing the sparse activation phenomenon, thereby expanding the representation space of these values and boosting the model's overall representational capacity. This is the key reason why our approach outperforms traditional dense models. To demonstrate how our method expands the activation value representation space, we apply t-SNE dimensionality reduction to the activation values of both the traditional dense model and the model trained using our method, as shown in Figure \ref{fig5}.

% 我们选取了500个中频词的激活值表示进行降维。具体来说，我们的方法采取2层专家子层和每个子层8个专家的设置，然后我们将第一个子层和第二个子层不同的专家的激活值连接到一起进行降维，这样可以保证我们的样本点与baseline模型的样本点数量一致。从图中可以看出，我们的方法在不同层的激活值覆盖的表示空间更广。这也是我们的方法比传统dense模型表现更好的直接原因。

% We select the activation representations of 500 mid-frequency words for dimensionality reduction. Specifically, our method utilizes a setup with two expert sub-layers, each containing eight experts. To ensure a fair comparison, we concatenate the activation values from different experts in the first and second sub-layers before performing dimensionality reduction, keeping the number of sample points consistent with the baseline model. As shown in the figure, our method covers a broader representation space across different layers, which directly explains its superior performance compared to traditional dense models. % TODO 这里应该说明一下broader为什么是更好的，是因为对样本更加可分吗？making samples more discriminative之类的？
% lmx：As shown in the figure, our method covers a broader representation space across different layers, making the token representations more discriminative and better separated. This enhanced separation in the representation space indicates that our model can better distinguish between different semantic concepts and capture more nuanced relationships between tokens, which directly explains its superior performance compared to traditional dense models. 下面是直接给你改上的：
We select the activation representations of 500 mid-frequency words for dimensionality reduction. Specifically, our method utilizes a setup with two expert sub-layers, each containing eight experts. To ensure a fair comparison, we concatenate the activation values from different experts in the first and second sub-layers before performing dimensionality reduction, keeping the number of sample points consistent with the baseline model. As shown in the figure, our method covers a broader representation space across different layers, making the token representations more discriminative and better separated. This enhanced separation in the representation space indicates that our model can better distinguish between different semantic concepts and capture more nuanced relationships between tokens, which directly explains its superior performance compared to traditional dense models.

\subsection{Mix Ratios of Different Pre-training Datasets}
\label{sec:appendix A.3}

% 参考其他开源模型技术报告和我们收集的各领域数据大小，我们最终将数据配比确定为如表1所示。
Referring to technical reports from other open-source models and the dataset sizes we collected from various domains, we finalized the data mixing ratios for each domain, as shown in Table \ref{tab:table1}.

\begin{table}[h]
\centering
\resizebox{0.25\textwidth}{!}{%
\begin{tabular}{ll}
\toprule
\textbf{Domain} & \textbf{Ratio} \\ \midrule
Cosmopedia      & 3.18\%         \\
Fineweb-Edu     & 86.31\%        \\
OpenWebMath     & 1.38\%         \\
StarCoder       & 9.13\%         \\ \bottomrule
\end{tabular}
}
\caption{Mixing ratios of pre-training data across different domains.}
\label{tab:table1}
\end{table}

\subsection{Evaluation Benchmarks}
\label{sec:appendix A.4}

To comprehensively assess the performance of our method, we evaluate across a diverse set of benchmarks covering various aspects. These benchmarks include tasks related to reading comprehension, language understanding, commonsense reasoning, and closed-book question answering.

\begin{itemize}
  \item Reading comprehension: We evaluate our method on SQuAD V2 \cite{rajpurkar2018know}, which tests the ability to answer questions based on given passages.
  \item Language understanding: We use LAMBADA \cite{paperno2016lambada}, a benchmark that requires models to predict the final word of a sentence, assessing long-range context understanding.
  \item Commonsense reasoning: We include ARC-Challenge \cite{DBLP:journals/corr/abs-1803-05457}, HellaSwag \cite{zellers2019hellaswag}, PIQA \cite{bisk2020piqa}, SIQA \cite{sap2019social}, and Winogrande \cite{sakaguchi2020winogrande}, which test the model’s ability to infer commonsense knowledge across various scenarios.
  \item Closed-book question answering: We assess factual knowledge recall using Natural Questions \cite{kwiatkowski2019natural} and TriviaQA \cite{joshi2017triviaqa}, where models must generate correct answers without relying on external documents.
\end{itemize}

\subsection{Training Configuration}
\label{sec:appendix A.5}

\begin{table}[h]
\centering
\resizebox{0.45\textwidth}{!}{%
\begin{tabular}{lccc}
\toprule
                           & \multicolumn{1}{c}{\textbf{Small}} & \multicolumn{1}{c}{\textbf{Medium}} & \multicolumn{1}{c}{\textbf{Large}} \\ \midrule
\textbf{Hidden Size}       & 1024                               & 2048                                & 4096                               \\
\textbf{Intermediate Size} & 4096                               & 8192                                & 11008                              \\
\textbf{Attention Heads}   & 16                                 & 8                                   & 32                                 \\
\textbf{Layers}            & 24                                 & 16                                  & 32                                 \\
\textbf{Learning Rate}     & 3e-4                            & 3e-4                              & 3e-4                             \\
\textbf{Weight Decay}      & 0.1                                & 0.1                                 & 0.1                                \\
\textbf{RMSNorm Epsilon}   & 1e-05                              & 1e-05                               & 1e-05                              \\ \bottomrule
\end{tabular}
}
\caption{Experimental training configuration.}
\label{tab:table7}
\end{table}

\end{document}